\documentclass[Afour,sageh,times]{sagej}

\usepackage{moreverb,url}
\usepackage{nameref}
\usepackage{color, colortbl}
\usepackage{xcolor}
\usepackage{csquotes}
\usepackage{bigstrut}
\usepackage{multirow}
\usepackage{balance}

\newcommand{\p}[1]{\smallskip \noindent \textbf{{#1}.}}
\newcommand\BibTeX{{\rmfamily B\kern-.05em \textsc{i\kern-.025em b}\kern-.08em
T\kern-.1667em\lower.7ex\hbox{E}\kern-.125emX}}

\usepackage{hyperref}
 \hypersetup{
     colorlinks=true,
     linkcolor=orange,
     filecolor=orange,
     citecolor=orange,      
     urlcolor=orange,
     }
\usepackage{perpage} %the perpage package
\MakePerPage{footnote} %the perpage package command 
\begin{document}

% %%%%%%%%%%%%%%%%%%%%%%%%%%%%%%%%%%%%%%%%%%%%%%%%%%%%%%%%%%%%%%%%%%%%%%%%%%%%%%%%%%%%%%%%

\runninghead{Habibian et al.}

\title{A Review of Communicating \\Robot Learning during \\Human-Robot Interaction}

\author{Soheil Habibian\affilnum{1}, Antonio Alvarez Valdivia\affilnum{2}, \\Laura H. Blumenschein\affilnum{2}, and Dylan P. Losey\affilnum{1}}

\affiliation{\affilnum{1}Virginia Tech, Department of Mechanical Engineering\\
\affilnum{2}Purdue University, Department of Mechanical Engineering}

\corrauth{Soheil Habibian, 
Department of Mechanical Engineering,
Virginia Tech, 
Blacksburg, VA,
24060, USA.}

\email{habibian@vt.edu}

%%%%%%%%%%%%%%%%%%%%%%%%%%%%%%%%%%%%%%%%%%%%%%%%%%%%%%%%%%%%%%%%%%%%%%%%%%%%%%%%%%%%%%%%

\begin{abstract}

For robots to seamlessly interact with humans, we first need to make sure that humans and robots understand one another.
Diverse algorithms have been developed to enable robots to learn from humans (i.e., transferring information from humans to robots).
In parallel, visual, haptic, and auditory communication interfaces have been designed to convey the robot's internal state to the human (i.e., transferring information from robots to humans).
Prior research often \textit{separates} these two directions of information transfer, and focuses primarily on either learning algorithms or communication interfaces.
By contrast, in this review we take an interdisciplinary approach to identify common themes and emerging trends that \textit{close the loop} between learning and communication.
Specifically, we survey state-of-the-art methods and outcomes for communicating a robot's learning back to the human teacher during human-robot interaction.
This discussion connects human-in-the-loop learning methods and explainable robot learning with multimodal feedback systems and measures of human-robot interaction.
We find that --- when learning and communication are developed together --- the resulting closed-loop system can lead to improved human teaching, increased human trust, and human-robot co-adaptation.
The paper includes a perspective on several of the interdisciplinary research themes and open questions that could advance how future robots communicate their learning to everyday operators.
Finally, we implement a selection of the reviewed methods in a case study where participants kinesthetically teach a robot arm.
This case study documents and tests an integrated approach for learning in ways that can be communicated, conveying this learning across multimodal interfaces, and measuring the resulting changes in human and robot behavior.

\end{abstract}


\maketitle

%%%%%%%%%%%%%%%%%%%%%%%%%%%%%%%%%%%%%%%%%%%%%%%%%%%%%%%%%%%%%%%%%%%%%%%%%%%%%%%%%%%%%%%%

\section{Introduction}

\begin{figure*}
\centering
\includegraphics[width=2\columnwidth]{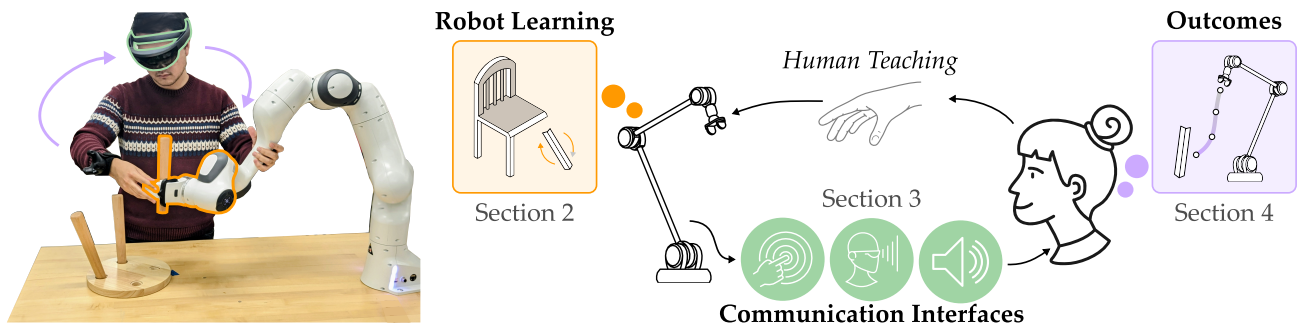}
\caption{Closing the loop with learning and communication during human-robot interaction. (Left) Example problem setting where a human teaches the robot arm to assemble chairs. (Right) Outline of our review paper. In Section~\ref{sec:learning} we first survey how robots can learn from humans in ways that can be communicated. As the robot uses these methods to determine what to communicate, it must then decide how to convey that information. In Section~\ref{sec:interfaces} we explore haptic, visual, and auditory interfaces for conveying a robot's latent state. The resulting communication closes the loop, and helps the human teacher understand what the robot has and has not learned. In Section~\ref{sec:models} we study how this feedback impacts the human teacher and the overall human-robot team.} \label{fig:front}
\end{figure*}

The robots we develop to interact with humans are becoming more intelligent, autonomous, and adaptable.
With these increases in complexity robots are no longer expected to just execute a programmed motion; instead, today's robots learn and extrapolate behaviors from human users.
For example, consider a human teaching a robot arm to assemble a chair (see Figure~\ref{fig:front}).
In the past, the robot arm acted as a tool that recorded the human's demonstration, and then replayed that motion with at most slight modifications \citep{brock2002elastic}.
But decades of advances in learning and control now enable more complex responses: we expect robots to extract the desired task (e.g. chair assembly), and then autonomously carry out that task in new contexts \citep{osa2018algorithmic, ravichandar2020recent}. 
With these modern tools, our example intelligent robot is expected to learn to add the remaining legs and complete the chair assembly after collecting demonstrations on how to insert a single leg.

Although this increased intelligence improves robot capabilities, it also obscures the robot's intent from everyday humans.
Consider our example of teaching a robot how to assemble a chair: as the human teacher provides demonstrations, they do not know what the robot has learned (i.e., has the robot learned to correctly insert chair legs?) or how the robot will behave (i.e., will the robot accidentally break chair legs when deployed?).
Learning robots are no longer predictable tools that always react to humans in the same way.
Instead, as contexts and inputs change, learning robots must output a variety of different behaviors that they were never explicitly shown.
This leaves a significant gap between (a) what the robot has learned and (b) what the human \textit{thinks} the robot has learned.
Repeatedly deploying the robot and observing its behavior can show parts of the robot's learning.
But for holistic, real-time understanding, we seek systems that directly close the gap between robot learners and human teachers.

In this review paper we survey works that attempt to \textit{close the loop} and communicate robot learning back to the human teacher during human-robot interaction. 
There have been recent advances in learning from humans --- including learning from demonstrations, human-in-the-loop reinforcement learning, and interactive imitation learning --- that extract autonomous robot behaviors through human-robot interaction.
Parallel research in haptics, soft robotics, augmented reality, and auditory interfaces has progressed fundamental knowledge of how robots can communicate latent information.
Importantly, these two branches of prior work are often \textit{separated}.
Work that seeks to explain robot learning primarily focuses on converting a robot's black-box models into interpretable feedback, and does not typically consider the physical interfaces used to convey that feedback to the human.
On the other hand, work on communication paradigms and multimodal interfaces improves our understanding of what types of signals a human can interpret, but these interfaces are not often designed with communicating robot learning in mind. 
Instead of treating interpretable robot learning and communication interfaces as separate topics, this review paper seeks to connect learning and communication into an interdisciplinary framework.

\p{Organization} Our overarching motivation is to connect recent trends at the intersection of robot learning and communication interfaces (see Figure~\ref{fig:front}). In \textbf{Section~\ref{sec:learning}} we review robot learning algorithms that implicitly or explicitly provide feedback to the human. Next, in \textbf{Section~\ref{sec:interfaces}} we explain how the community has developed communication interfaces to convey information known by the robot back to human users. Finally, in \textbf{Section~\ref{sec:models}} we explore the impact of closing the loop by surveying approaches that measure the resulting human-robot interaction and assess the outcomes in human understanding and robot learning.

As we discuss each area of research we will identify its relevant themes and directions. 
In \textbf{Section~\ref{sec:problems}} we then unify those trends to propose a set of open questions that should be answered to reach robots that seamlessly and intuitively reveal their learning back to human partners. 
We conclude our review with a case study in \textbf{Section~\ref{sec:userstudy}}. 
In this case study we test one integrated approach that learns from physical demonstrations in ways that can be communicated, leverages multimodal signals to convey the robot's learning in real-time, and then measures the changes in the human teaching and robot learning that are caused by the system's feedback\footnote{See user study videos here: \url{https://youtu.be/EXfQctqFzWs} \\
and our code repository here: \url{https://github.com/VT-Collab/communicating-robot-learning}}.

\p{Contributions} The primary contributions of our review article are as follows:

\p{Closing the Learning and Communication Loop} In Sections~\ref{sec:learning}, \ref{sec:interfaces}, and \ref{sec:models} we survey recent literature within human-robot interaction. 
We summarize learning architectures that bring the human into the learning process and implicitly or explicitly communicate what the robot has learned. 
We then overview communication interfaces that use visual, haptic, and/or auditory signals to intuitively and immersively convey a robot's latent information. 
Throughout this survey we connect works focused on learning and communication.
Finally, we review the outcomes of closing the loop, and summarize the measurement tools that can be used to assess how the human and robot perform and how the human builds a mental model of the robot learner.

\p{Identifying Research Trends} We organize Sections~\ref{sec:learning}, \ref{sec:interfaces}, and \ref{sec:models} to highlight the key themes that progress towards seamless communication of robot learning.
These themes include i) actively involving the human in the learning process, ii) converting learning models into interpretable signals through pre- or post-processing, iii) moving from visual to immersive, non-visual feedback, iv) incorporating multi-modal signals into interfaces to capture different aspects of the robot's learning, and v) measuring the human's understanding of the robot learner.
We group related works into each of these trends to illustrate the broader research directions that are currently being explored.

\p{Introducing Open Questions} We build on recent trends to identify a set of questions that must be addressed before we reach robots that seamlessly convey their learning to human partners. 
In Section~\ref{sec:problems} we introduce what we believe are the leading challenge areas: 
i) identifying representations of robot learning that are intuitive, interpretable, and comprehensive, 
ii) designing interfaces specifically for communicating robot learning, 
and iii) measuring the human's functional understanding of the robot learner in real time.
We explain how addressing each of these challenges will advance the community towards closing the loop between learning and communication.

\p{Conducting a Case Study} To demonstrate the core concepts of this review article, in Section~\ref{sec:userstudy} we perform a user study where participants kinesthetically teach a robot arm. We implement example algorithms and hardware to learn from the human's demonstrations, render the robot's learning on communication interfaces, and then measure the resulting effects on human-robot coordination. 
The methods from this case study are disseminated online so that other researchers can replicate and build upon our procedure.
Our results support the underlying hypothesis that closing the loop and communicating robot learning back to human teachers improves the overall interaction from both human and robot perspectives.
\begin{figure}[t!]
\centering
\includegraphics[width=1\columnwidth]{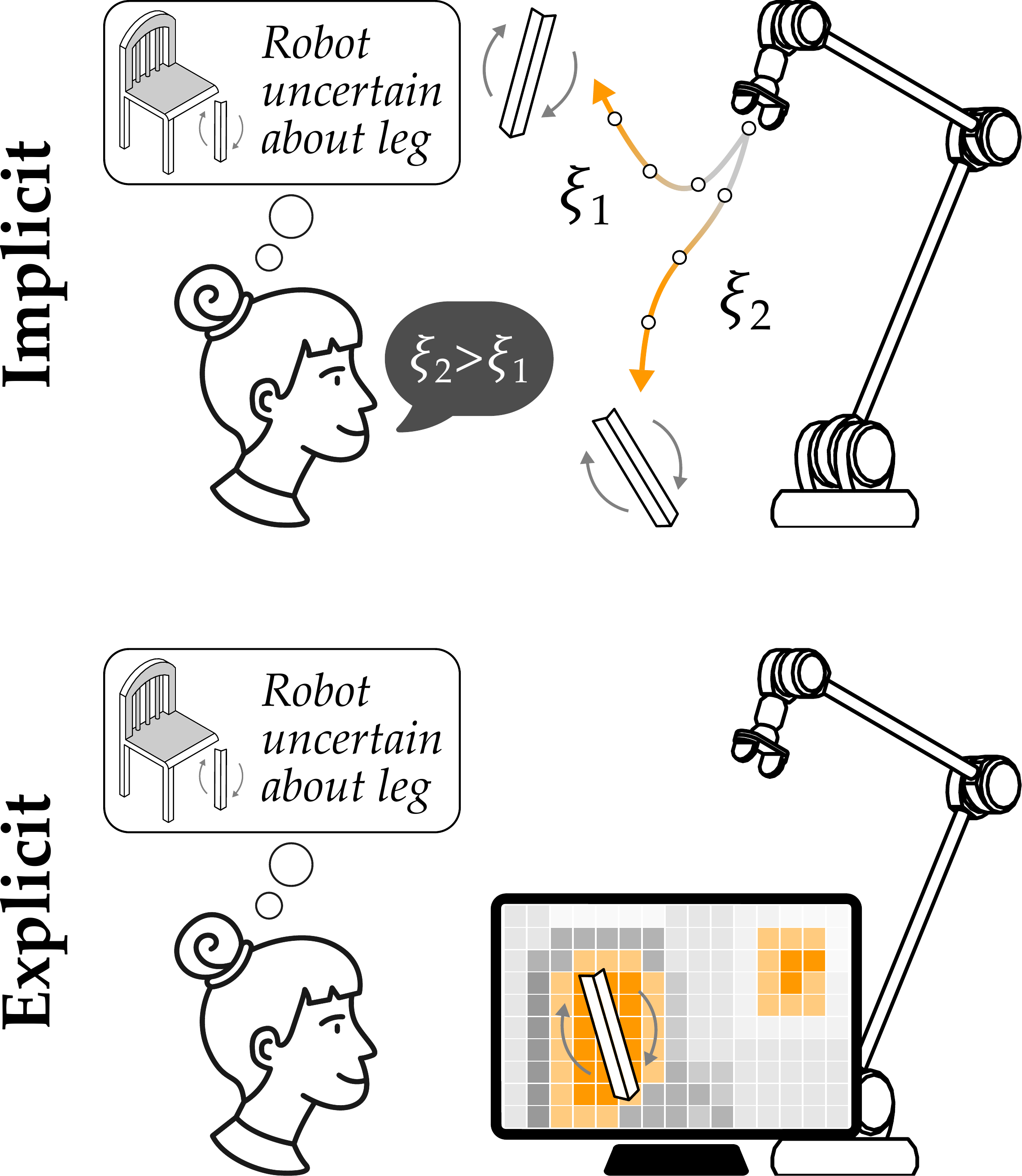}
\caption{Learning from humans in ways that can be communicated. (Top, Section~\ref{sec:L1}) Human-in-the-loop frameworks involve the human within an iterative learning process. At each iteration the robot shows behaviors that it has learned, and the human teacher provides labels. The robot \textit{implicitly} communicates its learning through the shown behaviors: here the robot's two trajectories suggest that the robot is uncertain about where to place the chair leg. 
(Bottom, Section~\ref{sec:L2}) Explainable learning frameworks are designed to intentionally communicate the robot's learning. The robot pre- or post-processes its learning architecture to extract an intuitive and human-friendly signal. The robot then \textit{explicitly} communicates its learning through this signal: here the robot highlights two regions on the screen it has learned to reach.} \label{fig:learning}
\end{figure}

\section{Learning from Humans in Ways \newline that can be Communicated} \label{sec:learning}

Human-robot interaction provides robots with an opportunity to learn tasks from a human teacher. This includes robots that ask the human questions, robots that imitate the human's behaviors, and robots that infer the human's objective. But while the robot learns, this learning process is often a \textit{black box} from the human's perspective \citep{hellstrom2018understandable}. In the worst case, the human cannot predict how the robot will behave --- or what the robot has learned to do --- until after the system is deployed and extensively tested.

In this section we survey works that learn tasks from humans in ways that facilitate communication back to the human teacher. 
We identify two key trends across these methods (see Figure~\ref{fig:learning}).
First, one set of learning approaches enables \textit{implicit} communication through the structure of their human-in-the-loop learning protocols (Section~\ref{sec:L1}).
These methods follow an interactive procedure where the robot shows behaviors and trajectories it has learned to the human at each iteration.
For example, here a robot arm might collect human demonstrations of how to carry a chair leg, start to execute its learned behavior in the environment, and then stop and ask the human for additional guidance when it is unsure.
By observing snippets of the robot's learned behavior throughout the learning process the human teacher continually develops an understanding of what the robot does and does not know.

Second, another set of learning approaches enables \textit{explicit} communication by designing their learning networks to extract interpretable and intuitive representations (Section~\ref{sec:L2}).
These methods learn from human data, and then pre- or post-process their learned models to recover a user-friendly signal that captures what the robot has learned.
For example, here a robot arm might collect human demonstrations of how to carry a chair leg, extract the system states that have the largest impact on task performance, and then highlight those critical states on a visual interface for the human.
Within this group of works the human infers what the robot has and has not learned by reasoning over the robot's explicit feedback signals.
Our review of these explicit methods is intended to be complementary to existing surveys on explainable artificial intelligence \citep{silva2023explainable}. 
However, here we specifically focus on algorithms which extract learning signals that can be displayed on communication interfaces during human-robot interaction.

We note that these \textit{implicit} and \textit{explicit} trends are not mutually exclusive.
Robots can follow interactive learning protocols that show learned behaviors at each iteration, and then process their results to obtain explicit feedback signals that summarize their learned models.

% Section 1
\subsection{Implicitly Communicating Learning: Human-in-the-Loop Frameworks} \label{sec:L1}

We start with interactive learning frameworks that involve the human in the learning process.
This includes human-in-the-loop reinforcement learning, active reward learning, interactive imitation learning, and learning from corrections. 
These approaches incorporate the human teacher within an iterative protocol that alternates between the human teaching the robot and the robot showing the behaviors it has learned.
The important theme here is that --- because the human is in the learning loop --- \textit{implicit} communication occurs when the human observes how the robot learns and improves over time.
We recognize that communicating robot learning may not be the primary intent of these methods; indeed, the surveyed papers often focus on efficient learning from the robot's perspective.
However, we will review how communicating the robot's learning emerges as a side-effect of these interacting learning frameworks, and how different learning approaches have augmented this communication.

%%%%%% Reinforcement Learning
\p{Reinforcement Learning} In reinforcement learning (RL) the robot uses trial and error to find a policy that maximizes its reward function. Traditionally this reward function is a mathematical expression given to the robot \citep{kober2013reinforcement}. For example, the reward for an autonomous car could be inversely proportional to the time it takes for that car to reach its destination. But often it is challenging to mathematically formulate the right reward function --- e.g., how do we write equations for safe driving? --- especially when different users may want the robot to optimize for different rewards --- e.g., an aggressive driver vs. a defensive driver. \textit{Human-in-the-loop RL} brings a human teacher into the learning process to address these concerns. Specifically, in human-in-the-loop RL the robot relies on the human to provide the reward signal: during each iteration the robot shows its learned behaviors to the human, and the human expert assigns rewards to these behaviors.

For example, in \cite{warnell2018deep} the robot alternates between training on its own and showing its learned policy to the human. After each round of training the robot demonstrates its behavior in a simulated environment, and the human teacher assigns a scalar reward value to indicate the quality of the robot's motion. For instance, the user might assign $0$ reward if the robot collides with an obstacle and $1$ reward if the robot successfully completes its task. From this human feedback the robot iteratively extrapolates a model of the human's reward function and then retrains to optimize that estimated reward. The human's inputs guide the RL process \citep{celemin2019fast}. \cite{reddy2018shared} apply a similar approach to shared autonomy settings where both the human and the robot have control over the robot's actions (e.g., a human teleoperating an autonomous drone). Here the human works with the robot throughout the task and then provides reward feedback at the end of each interaction to indicate if the human-robot team has completed that task successfully. \cite{meng2020learning} learn behaviors that enhance interaction by using the human's occupancy (i.e., the time the human spends near the robot) as the human-provided reward signal. The resulting robot behavior increases engagement across multiple human users in a public setting.

These RL approaches bring the human into an interactive learning process: the robot learns from the reward labels the human assigns, and the human gets a sense of what the robot has learned by observing the behaviors they are asked to label.
However, there are still two challenges from the human's perspective.
\textit{First}, it is difficult for users to consistently rate robot behaviors on an absolute spectrum (e.g., scoring the robot's motions between $0$ and $1$). 
Works such as \cite{lee2021pebble} and \cite{hejna2023few} address this problem by showing the human pairs of robot behaviors, and then asking the human to select the one that better matches their own reward function.
\textit{Second}, if the human needs to provide a reward signal at every learning loop, the overall RL process quickly becomes too time consuming for practical use.
Methods including \cite{hejna2023few}, \cite{xie2022ask}, and \cite{warnell2018deep} purposefully reduce the amount of human interaction by intelligently selecting the pairs of robot behaviors or only asking the human to provide a reward label when the robot is uncertain.
While they are focused on the learning, both of these human-centered innovations augment the implicit communication during human-in-the-loop RL.
Showing the human pairs of robot behaviors can make it easier for humans to assess the robot's learning progress, and curating these behaviors to focus on regions of uncertainty can better align the robot's implicit communication with its learning.

%%%%%% Active Learning
\p{Active Learning} Another class of learning methods that show behaviors and then ask for feedback is \textit{active preference-based reward learning}. Within active learning the robot asks the human multiple choice questions --- e.g., a robot arm demonstrates different ways to carry a chair leg --- and the human responds by picking their favorite option (i.e., the behavior that best aligns with their preferences). A na\"ive robot might ask questions completely at random. But in active learning the robot intentionally selects questions that will efficiently gather information from the human \citep{cakmak2012designing, sadighactive, biyik2022aprel, quintero2022human}. Specifically, in active preference-based reward learning the robot designs queries to rapidly infer the human's reward function, which it can then optimize to complete the desired task. The robot must be particularly careful with the questions it asks to avoid learning --- and optimizing for --- the wrong reward \citep{tien2022causal}.

During active learning the human teacher gets implicit feedback about what the robot has learned based on the multiple-choice questions that the robot asks. 
Consider the example at the top of Figure~\ref{fig:learning}.
If the robot shows one trajectory that carries the chair leg away from the table, $\xi_1$, and another trajectory that moves towards the table, $\xi_2$, the human might infer that the robot arm is uncertain about where to place the chair leg.
Related works have augmented this implicit communication within active learning by accounting for the human teacher.
For instance, \cite{biyik2019asking} and \cite{bullard2019active} purposely ask questions that are easy for the human to answer.
To accomplish this, these works encourage the robot to display trajectories with distinct, intuitive differences --- making it easier for the human to spot the differences and select their preference.
\cite{habibian2022here} take this one step further by selecting questions that align with what the robot does and does not know.
Returning to our chair example in Figure~\ref{fig:learning}, imagine that the robot has learned to grasp the leg but is not sure where to place it.
Under \cite{habibian2022here} the robot might select a question with two trajectories, where both trajectories reach the leg (conveying what the robot knows) and then move towards different locations (conveying what the robot is uncertain about).

\cite{tucker2020preference} demonstrate a practical application of this implicit communication for assistive lower-limb exoskeletons.
Lower-limb exoskeletons need to identify their user's preferred walking gait, and the correct gait can vary from one user to another.
Within \cite{tucker2020preference} the authors take an active learning approach: the robot selects a set of gait parameters, the participants try walking with each chosen gait, and then the robot updates the gait options based on the participant's feedback.
Unlike the prior works we have reviewed --- where the robot's behaviors are watched by the human --- here the human can physically feel how the robot is learning and adapting to their preferences over time.

%%%%%% Imitation learning
\p{Imitation Learning} In human-in-the-loop RL and active learning the robot attempts to infer the human's reward function. By contrast, during imitation learning the robot directly learns a control policy (i.e., a mapping from states to actions) based on demonstrations provided by the human teacher. 
In accordance with our trend on the implicit communication that results from putting a human in the learning loop, here we specifically focus on \textit{interactive imitation learning} \citep{celemin2022interactive}. Imagine a user teaching a robot arm to pick up objects and place them on a table \citep{mandlekar2020human}. At first the robot does not know the desired task, and the user must kinesthetically guide the robot through a complete demonstration of reaching for objects, grasping them, and then sorting them on the table. As the robot learns from these demonstrations it begins to imitate the human and perform parts of the task autonomously. But what happens when the robot makes a mistake or encounters a new object it has never seen before? Interactive imitation learning enables the human to intervene online --- as the robot performs the task --- and provide additional guidance that the robot can learn from to improve its future iterations. Each time the human intervenes they label robot states with the correct actions (e.g., the human demonstrates snippets of the ideal policy).

Here implicit communication occurs when the human watches the robot and intervenes to correct its actions.
For example, if the human notices that the robot arm has sorted an object incorrectly, the human might infer the robot still needs to learn that part of its policy.
A key challenge for interactive imitation learning is determining \textit{when} the human should intervene: at a given state, should the robot try to act autonomously or ask the human for guidance?
Increased human involvement is an opportunity for additional implicit feedback about the robot's learning, but at the cost of requiring more human attention and effort.
Under \textit{human-gated} approaches such as \cite{kelly2019hg} and \cite{mandlekar2020human} the robot relies on the human to determine when to intervene. 
By default the robot executes its current learned policy; when the human chooses to intervene and provides new state-action pairs, the robot retrains its policy to match the human's demonstrations.
The robot then follows this updated policy during future iterations.
Alternatively, in \textit{robot-gated} interactive imitation learning the robot decides when to prompt the human for additional guidance \citep{hoque2021thriftydagger}.
This reduces the human's burden --- because the human does not need to always monitor the robot's behavior --- but also means that the robot must keep track of what it does not know.
For instance, in \cite{hoque2021thriftydagger} the robot maintains an ensemble of learned policies, and queries the human when these models diverge (i.e., when the robot is uncertain about the correct action).
Regardless of whether we leverage human- or robot-gated approaches, interactive imitation learning provides implicit feedback when the robot makes a mistake and needs human guidance.

Interestingly, the human can also gather implicit feedback from their own interventions.
Once the human demonstrates how the robot should behave in a specific scenario, the human naturally expects the intelligent robot to understand that scenario moving forward.
Put another way, the human might assume the robot will not make the same mistake twice.
Methods like \cite{spencer2022expert} and \cite{chisari2022correct} support this assumption by updating the robot's policy in the entire region around each intervention to better align with the human's desired behavior.
This results in robots that are less likely to make the same mistake again --- and thus the human can infer the robot knows how to behave in regions where they have previously provided guidance.

%%%%%% Learning from Corrections
\p{Corrections} Building on interactive imitation learning, a final paradigm that implicitly communicates with the human is \textit{corrections}.
Corrections are different from interactive imitation learning in two ways:
i) corrections are often physical, where the user kinethetically modifies the motion of their robot, and 
ii) instead of directly updating a policy, the robot uses corrections to learn the reward function that it should optimize.
Imagine a robot arm carrying a chair leg.
During corrections the robot starts to execute its planned motion, and the human can kinesthetically push, pull, and guide the robot to correct its motion \citep{haddadin2016physical}. 
The robot incorporates these corrections to learn a reward function in real-time: the robot changes its autonomous behavior to finish the task correctly during the current iteration, and also remembers that correction for future iterations.
In practice, corrections provide the human an opportunity to refine and fine-tune the robot's behavior.

Physical corrections open a \textit{tactile} communication channel between the human teacher and robot learner \citep{kronander2013learning, rozo2016learning}.
When a human physically teaches the robot arm, they can perceive the forces and torques that the robot uses to resist or align with their corrections.
For example, if the human is trying to guide the robot arm closer to the table --- and they feel the robot continually pushing back against their correction --- the human might infer that the robot does not understand the desired task.
Works such as \cite{losey2022physical}, \cite{jin2022learning}, and \cite{jain2015learning} consider the human's perspective during physical corrections.
These methods recognize that it is difficult to provide perfect corrections: robot arms are high-dimensional, and it is hard for humans to precisely orchestrate all the joints of these arms to indicate an exact motion.
As a result, \cite{losey2022physical}, \cite{jin2022learning}, and \cite{jain2015learning} treat the human's correction as an incremental improvement over the robot's current behavior.
Multi-modal implicit communication occurs as the human iteratively makes these corrections.
When the human sees that the robot is making a mistake (visual feedback), the human starts to kinesthetically guide the robot back towards their desired behavior. 
As the robot learns it updates its trajectory in real-time: the human can feel the changing forces and torques applied by the arm during the correction (kinesthetic feedback), and then see the change in the robot's autonomous motion after they let go of the arm.

The frequency of robot learning can also affect this implicit communication. 
In some approaches the robot collects a set of multiple corrections, and then learns once from this entire batch \citep{cui2018active, jain2015learning}.
In other approaches the robot performs online gradient descent, and updates its estimate in in real-time after each individual correction \citep{losey2022physical}.
Shared autonomy paradigms similarly place the human in the loop at every timestep, where each robot action is a combination of the human's input and the robot's autonomous assistance \citep{jain2019probabilistic, javdani2018shared, hagenow2021corrective}.
When the robot updates its behavior in batches the human gets fewer points of feedback about the robot's learning: it may not be clear to the user which correction(s) caused the robot to learn the task.
By contrast, when the human receives immediate feedback after each correction, they can observe how the robot interpreted their correction and altered its behavior.

\p{Summary} Interactive learning frameworks place the human within the robot's learning loop.
These methods do not use a physical communication interface (i.e., there is not a visual, auditory, or haptic interface for conveying the robot's learning).
Instead, the human gets implicit feedback about what the robot has learned by observing the changes in robot behavior over time.
By providing reward signals, answering queries, demonstrating trajectory snippets, and physically correcting the robot's behavior, the human implicitly gathers information about what parts of the task the robot has learned and which parts it is still confused about.

% Section 2
\subsection{Explicitly Communicating Learning: Pre-Hoc and Post-Hoc Frameworks} \label{sec:L2}

In Section~\ref{sec:L1} we reviewed human-in-the-loop paradigms that provide implicit feedback about the robot's learning. 
Communication is often an auxiliary outcome of these methods: when the robot displays behaviors to the human teacher, the human has a chance to observe snippets of what the robot has and has not learned.
By contrast, in this section we will survey learning frameworks that are \textit{explicitly} designed to communicate the robot's learning.
These methods extract human-friendly signals that summarize what the robot has learned; e.g., a sentence explaining the robot's decision making or an image showing where the robot is confused.
Consistent with related surveys, we will divide explicit methods into two groups: \textit{pre-hoc} and \textit{post-hoc} \citep{milani2022survey}.
Pre-hoc approaches design the robot's learning architecture so that this architecture itself is easy for humans to parse and explain.
Post-hoc approaches learn using existing methods, and then convert the learned models back into feedback signals for the human.

%%%%%% intrinsic explainable policies
\subsubsection{Pre-Hoc Frameworks} 

Pre-hoc approaches assemble the robot's learning models out of intuitive building blocks.
By looking at these building blocks, the human can directly grasp what the robot has learned.
For example, imagine a robot arm learning how to open a door.
Instead of learning a single model of the overall task, the robot might learn a sequence of waypoints: reaching for the door, turning the handle, and pulling the door open.
By observing each learned waypoint --- and seeing the errors in those waypoints --- the human can identify what parts of the task the robot knows and where the robot is likely to fail.
More generally, explicit pre-hoc approaches take advantage of behavior trees, hierarchies, and wrapper models.

%%%%%% behavior / decision trees
\p{Behavior and Decision Trees} 
Tree-based approaches organize the robot's learning into a flowchart-like structure.
The nodes of the tree capture subtasks or key decisions, and the edges show how one decision might lead to another. 
For instance, when an autonomous car approaches a light, its decision tree could include a node that chooses to go if the light is green, and its edges might lead to other nodes that decide which way the autonomous car will turn.
Human users can read through these tree structures to explain the robot's behavior; e.g., to understand why the autonomous car stops or turns right.
\cite{han2021building} build upon these ideas to create behavior trees that explain the robot's decision making in terms of goals, subgoals, steps, and actions.
Users can query this robot to get explicit communication about the robot's learning (i.e., the robot can answer questions like ``Why are you doing this?").
In \cite{french2019learning} the authors use imitation learning to extract a behavior tree from human demonstrations.
After humans demonstrate the task for the robot, they can then refer to the decision tree to see how the robot interpreted their demonstrations --- if any of the nodes or edges are incorrect, the human can directly fix those components.
Tree-based models have similarly been used within reinforcement learning \citep{paleja2022learning, li2019formal} to structure the robot's policy in a way that human users can parse and understand.

%%%%%% hieraarchical structure
\p{Hierarchies} 
Other pre-hoc approaches split the agent's behavior into low-level waypoints (or subtasks) that are individually interpretable. 
The human can check the robot's overall learning process by monitoring each of the subtasks.
Here we refer back to our example of a robot arm learning how to open the door: instead of reasoning over the robot's mapping from states to actions, it can help users to think in terms of subtasks like reaching for the door, turning the handle, and pulling the door open.  
\cite{beyret2019dot} learn this hierarchy by combining two reinforcement learning agents: a high-level agent that determines the subtasks the robot arm should complete, and a low-level agent that learns how to move the robot's joints to complete each subtask.
The robot displays its learned subtasks to the human in a simulated graphical environment so the human can monitor the robot's progress.
Similarly, \cite{liu2018interactive} learn a hierarchical policy from human demonstrations, and then visualize that hierarchy in augmented reality. Human teachers wearing augmented reality displays can directly interact with the rendered hierarchy to check and modify the subtasks. In practice, hierarchical approaches help human teachers focus on the robot's high-level steps instead of the low-level processes the robot uses to move between steps.

%%%%%% wrapper models
\p{Wrappers} 
Wrapper models bridge the gap between pre-hoc structures and post-hoc representations. 
In wrapping approaches the robot first learns a policy using traditional neural networks (e.g., methods from Section~\ref{sec:L1}).
The robot then applies a wrapping algorithm to convert that complex, nonlinear function into a policy with an explainable, human-friendly structure that has been pre-defined.
For example, in \cite{bastani2018verifiable} the authors take a neural network learned with interactive imitation learning, and then search for a decision tree that best matches the behaviors of the neural network.
The resulting decision tree is a simplified version of the learned policy: although the original neural network may have captured more complex behaviors, the decision tree provides a human-friendly version that users can interpret.
Along the same lines, \cite{kenny2023towards} convert the original policy that the robot learns into a set of human-defined prototypes.
Each prototype is understandable to the human --- e.g., an autonomous car turning right or going straight --- and the wrapper forces the robot's learned behavior to be a combination of these known prototypes.
The user studies performed in \cite{kenny2023towards} suggest that this approach helps humans better predict the robot's behavior.
In \cite{kenny2023towards} the robot displayed its learned performance for each of the prototypes on a graphical user interface: after observing these short videos, participants were more accurately able to guess how the robot would behave at its current state.

%%%%%% post-hoc frameworks
\subsubsection{Post-Hoc Frameworks}

Pre-hoc methods structure their learning so that the models themselves are intuitive and easy to interpret.
By contrast, \textit{post-hoc} approaches do not change the learning algorithm. 
Instead, they apply post-processing techniques to extract an understandable feedback signal from the robot's learned models.
Consider our chair example in Figure~\ref{fig:learning}.
Here the robot might first collect human demonstrations, then learn a policy to match the human, and finally apply post-hoc methods to identify critical task states (e.g., goal locations).
The robot can visualize these critical states on a monitor for the human teacher; by observing this explicit feedback, the human infers what the robot has learned.
Post-hoc frameworks include saliency methods that highlight critical states, machine teaching approaches that convey the robot's objective, and natural language templates that articulate the robot's policy.

%%%%%% saliency techniques
\p{Saliency Methods} 
Saliency algorithms communicate the policy that a robot has learned by highlighting the robot's actions in specific states.
Consider Figure~\ref{fig:learning} where a human teacher is trying to determine whether their robot arm has learned to autonomously insert chair legs.
It would be too time consuming for the human to watch how the robot arm reaches for and carries the leg at every single state of the environment.
Instead, \cite{watkins2021explaining} and \cite{olson2021counterfactual} summarize the robot's policy by finding the \textit{critical states} where --- if the robot were to deviate from its learned policy --- the robot predicts it would achieve significantly lower long-term reward.
Returning to our chair example, critical states could occur when the robot arm is placing the leg: here deviating from the robot's policy (e.g., dropping the leg on the table instead of placing it in the chair base) could cause the task to fail and incur a large reward penalty.
The robot renders each of these critical states on a graphical user interface for the human to inspect.
By observing this small set of state-action pairs, the human gets a sense of whether the robot has learned correctly overall and is ready to be deployed \citep{watkins2021explaining, olson2021counterfactual}.
In practice, just seeing disconnected states and the actions in those states can make it challenging for humans to picture the robot's holistic behavior.
Accordingly, methods such as \cite{amir2018highlights} and \cite{du2023conveying} reveal trajectories that pass through as many of these critical states as possible, placing the robot's behavior at these states within a larger task context.
Related works have also projected lights or images onto real-world environments to focus the human's attention on critical states during human-robot interaction \citep{andersen2016projecting}.
We note a trade-off between --- on the one hand --- showing as many critical states as possible, and perhaps overloading the human with information, and --- on the other hand --- restricting the robot's feedback to a few critical states, which may not be sufficient to capture the robot's learned policy.

%%%%%% machine teaching
\p{Machine Teaching in Robotics}
Another post-hoc framework leverages \textit{machine teaching} to communicate the objective (e.g., the reward function) that the robot has learned.
Machine teaching --- also known as algorithmic teaching --- applies to settings where the robot has access to some hidden information, and the robot needs to select its own actions or demonstrations to convey this information to the human \citep{cakmak2012algorithmic, brown2019machine}.
For example, in \cite{huang2019enabling} the authors apply machine teaching to communicate the driving preferences an autonomous car has learned.
The autonomous car selects environments and trajectories to exhibit whether it is an aggressive or defensive driver; e.g., the autonomous car might play videos of it merging directly in front of another car to indicate that it is optimizing for aggressive behaviors.
A key challenge within machine teaching is determining how humans will interpret the robot's demonstrations.
One user might see videos of the autonomous car merging and infer that it is aggressive; another human might think that the autonomous car just likes to change lanes.
\cite{lee2021machine} try to make it easier for humans to interpret the robot's demonstrations by gradually increasing their complexity.
At first the robot shows simple behaviors (e.g., driving rapidly), and over repeated videos the robot reveals more complex trajectories (e.g., weaving through traffic).
As the community continues to develop more accurate models and measurements of how humans interpret robot behavior (Section~\ref{sec:H1}), machine teaching approaches can better tune the robot's demonstrations to align with those human models and communicate the robot's learned objective.

%%%%%% language-based techniques
\p{Explaining Policies with Natural Language}
One final type of post-hoc frameworks uses post-processing to translate the robot's learning into natural language utterances.
For example, \cite{hayes2017improving} give the robot a set of sentence templates, and then the robot automatically completes these templates based on its learned policy.
The human first asks the robot arm why it is taking an action (e.g., \textit{``when do you inspect a part?''}), and the robot fits that query to a pre-defined template (e.g., \textit{``when do you \{action\}''}).
The robot then searches for states that meet the query criteria, and finally determines logical combinations of communicable predicates to match its control policy (e.g., \textit{``I inspect a part when that part is the wrong size or shape''}).
Related work by \cite{brawer2023interactive} extends this communication into a bidirectional exchange.
Similar to \cite{hayes2017improving}, the robot can output sentences to explain the logic behind its decision making, but now the human can also respond to the robot and refine its run-time behavior.
When applied to the previous example, under \cite{brawer2023interactive} the human might tell the robot \textit{``do not inspect parts based on size,''} and the robot will temporarily modify its policy to match the human's directive.
In Section~\ref{sec:I1-2} we expand on how auditory interfaces can be used to convey natural language between human and robot.

% section summary
\p{Summary} 
Both pre-hoc and post-hoc frameworks offer a way for robots to explicitly communicate their learning to nearby humans.
These methods convert nonlinear, high-dimensional, and unintuitive neural networks into human-friendly signals: e.g., a sequence of waypoints, a video of critical states, or a sentence explaining the robot's policy.
We emphasize that these \textit{explicit} methods are complementary to the \textit{implicit} communication frameworks from Section~\ref{sec:L1}.
For instance, a robot can first apply human-in-the-loop algorithms to iteratively gather data from the human while showing its learned behaviors; the robot can then leverage post-hoc frameworks to convert the learned models into an explicit signal that summarizes what the robot has learned.
But while these implicit and explicit learning methods output signals to capture robot learning, it is still not clear how the robot should effectively convey these signals back to the human operator. 
Accordingly, in the next section we survey visual, haptic, and auditory interfaces that enable robots to communicate their latent information.

\begin{figure}[t!]
\centering
\includegraphics[width=1.0\columnwidth]{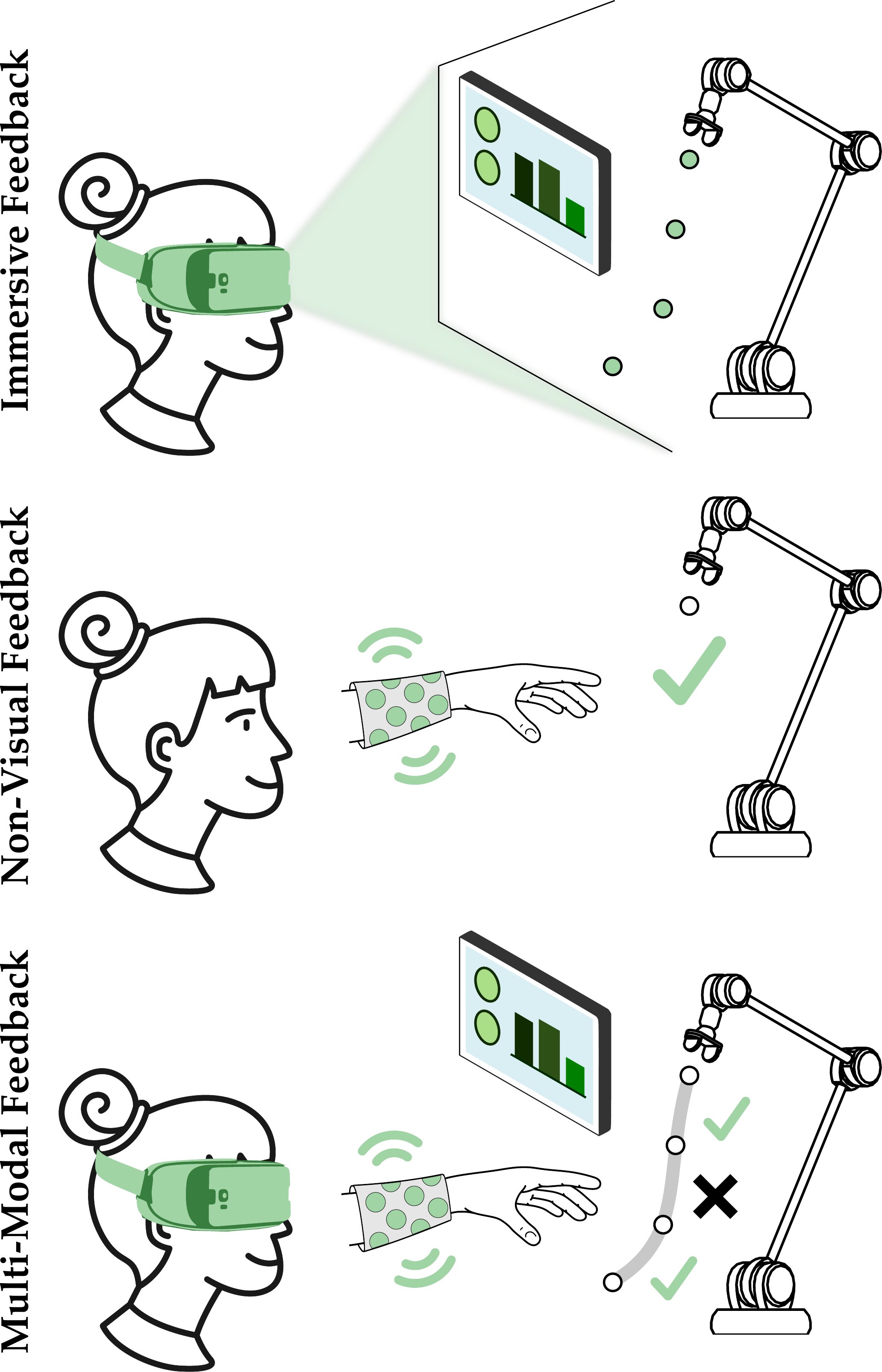}
\caption{Interfaces for communicating robot learning. The human previously provided demonstrations and is now observing feedback about what the robot has learned. (Top, Section~\ref{sec:I1-1}) Rather than presenting information on a computer monitor, recent works seek to provide more immersive and user-friendly visual feedback. Here the robot uses augmented reality to render virtual waypoints within its physical workspace. These waypoints indicate where the robot has learned to go. (Middle, Section~\ref{sec:I1-2}) Visual feedback can be intrusive or ineffective if the human is distracted. Other works leverage auditory and haptic interfaces to convey alternate representations of the robot's learning. For instance, a haptic wristband vibrates to notify the human that the robot is confident about the next waypoint.
(Bottom, Section~\ref{sec:I2}) Different interface modalities are able to convey different types of information. By combining multiple interfaces, the robot can provide more holistic feedback about what it has learned. In this scenario, the robot uses augmented reality to visualize its learned trajectory, while haptic cues indicate the parts of the trajectory where the robot is confident or confused.} \label{fig:interfaces}
\end{figure}

\section{Interfaces for Communicating \newline Robot Learning}\label{sec:interfaces}

In Section~\ref{sec:learning} we summarized how learning frameworks can be designed to produce implicit or explicit representations of robot learning.
This addresses the question of \textit{what} the robot should communicate; answering the question of \textit{how} the robot should communicate this data is a different problem. To answer the \textit{how}, we need to understand how interfaces can convert abstract learned information into tangible and intuitive feedback for human users. 
Interfaces provide a channel for communicating information from the robot to the human, determining what the human can perceive, identify, and ultimately comprehend about the robot's state. 
In this section we review the literature on interface design and development for human-robot interaction. Within our analysis we highlight two research trends: first, a trend from traditional visual displays towards more immersive interfaces using augmented reality, audio, or haptics, and second, a trend from single modality to multi-modality interfaces (e.g., from visual alone to visual plus auditory).

\subsection{Moving From Visual \newline to Non-Visual Feedback} \label{sec:I1}

Traditionally, human-robot systems have been developed with the assumption that humans are interacting with a screen. Screens are a convenient interface: they are readily available, easy to configure, and provide flexibility on the type of materials that are presented (e.g. text, images, videos, animations, and symbols).  
Given a problem setting that can be flattened to a 2D plane --- such as an inverted pendulum or some autonomous driving environments --- a screen can provide almost complete information about the simulated motion. This suffices for testing the transparency of learned behaviors \citep{paleja2022learning,beyret2019dot}. However, screens have difficulty accurately conveying the 3D world \citep{walker2023virtual}, and the limited field of view, lack of depth perception, and fixed orientation may further confuse users \citep{chen2007human}. This mismatch can lead to errors when humans use screens to interpret feedback from robot learners \citep{diehl2020augmented}. To improve the accurate transfer of information, work on conveying robot learning has begun to investigate communication methods which are \textit{immersed} in the physical space of the interaction \citep{reardon2018come, luria2017comparing, bolano2021deploying, chu2022multisensory}. For immersive visual feedback, researchers are exploiting the flexibility of virtual reality (VR) and augmented reality (AR) \citep{walker2023virtual}. In parallel, to better communicate non-visual information such as force, researchers are leveraging other sensing modalities like auditory and haptic feedback \citep{kassem2022happening}.
Here we will explore how the field of communication for robot learning is broadening its scope beyond screens and into these immersive interfaces.

\subsubsection{Towards More Immersive Visual Feedback} \label{sec:I1-1}

Recent works on human-robot interaction often use screens to provide information to humans \citep{rossi2021evaluation, shah2021jedai, pascher2023communicate, cleaver2021dynamic, aubert2018designing}. Looking specifically at communicating robot learning, several of the algorithms described in Section~\ref{sec:learning} visualize the robot's learning on a computer screen \citep{paleja2022learning, beyret2019dot, huang2019enabling, das2021explainable}. However, screens that are not part of the robotic system tend to pull focus away from the interaction and act as a middleman between the robot and human, leading to less immersion, less connection, and more distraction \citep{suzuki2022augmented, valdivia2023wrapping}. These screens also tend to poorly communicate information involving spatial components \citep{walker2023virtual}. Imagine a person teaching a robot arm to attach chair legs; the screen displays a simulated movement that the robot has learned from a single angle projected into a 2D representation. This flat visual representation potentially hides errors in the depth direction, and does not capture important non-visual features such as the force required to attach the leg. As such, options for information-rich feedback that allow users to stay focused on the task have been a major area of interface design.

\p{Augmented and Virtual Reality} AR and VR are the closest alternatives to traditional visual interfaces. These options offer many of the same features as screens, but in a more immersive format. 
AR enables interface designers to embed virtual elements into the real interaction space, adding a spatial component to the information \citep{wang2023explainable, zolotas2018head}, while VR allows wearers to interact within a completely virtual version of the environment \citep{reardon2018come, ye2023robot}. In the context of robot learning, researchers have utilized both augmented and virtual reality interfaces, enabling humans to observe robot motions either during training or in-between demonstrations \citep{coronado2020visual, diehl2020augmented, liu2018interactive, wang2023explainable}. \cite{diehl2020augmented} compare AR to traditional visual displays by showing simulated trajectories in both interfaces and asking if the trajectories represent correct or erroneous examples. Both approaches (augmented reality and the screen) had the same error rates for users identifying erroneous behaviors. However, participants preferred AR because it seemed ``more natural'' and usable. \cite{liu2018interactive} apply augmented reality to display the robot's learned decision tree and to allow users to modify that tree. Here AR provides a teaching method in addition to the communication mechanism, so that users are able to seamlessly transition between viewing the learned policy and selecting portions of that policy to update with new demonstrations. Another example where augmented reality serves as both interface and input device is \cite{wang2023explainable}, which focuses on the reactions of novice users to feedback. This results in some unique applications of augmented reality as a communication interface; for instance, objects within the space are rendered with AR labels so that the user knows what the robot observes. These findings suggest that AR can provide equivalent communication to screens, and that users perceive the interfaces as less obtrusive and more natural since the information is overlaid onto the world.

Other recent examples of AR and VR in human-robot interaction attempt more broadly to communicate spatially grounded 3D information. A wide range of AR/VR interfaces have been created to convey motion intent for robot arms \citep{rosen2019communicating} and drones \citep{walker2018communicating}. \cite{reardon2018come} developed an AR interface that allows the robot to share its latent information with a human teammate using a machine teaching approach. Similarly, \cite{tabrez2022descriptive} demonstrated that AR can be used to simultaneously communicate concrete information --- such as planned motions --- alongside abstract visualizations --- such as icons representing decision making. Overall, these results show AR and VR's evolving capabilities for revealing the robot's latent state in more general human-robot interaction settings.

\p{Lights and Projections} While not as capable of the rich visual detail found in AR and VR, some researchers have sought to streamline visual information into binary or discretized options. Lights --- and other visual additions to the robot itself --- minimize the potential for confusion when communicating robot status and intent \citep{demarco2014underwater}. Within robot learning, communicating the robot's motions in a given situation (i.e., its intent) and recognizing the robot's need for help (i.e., its status) are important to understand the current learning state. \cite{tang2019development} develop a skin of LED lights wrapped around the robot arm.
By observing this skin, human teachers can quickly and easily monitor the robot's intention and status. This minimal, yet easy-to-interpret, communication allows a single user to monitor and teach multiple robots. 
Lights have also been modulated in their rhythm \citep{demarco2014underwater} and color \citep{song2019designing, koay2014social} to communicate the robot's status and intent.

One step in complexity above basic lights are projections.
Here the robot projects simple images onto itself or nearby surfaces. Concrete information, such as heading direction, can be symbolically displayed using arrows and pointers \citep{shrestha2018communicating, watanabe2015communicating}. Projections have also been used to communicate less concrete concepts, such as robot intention, by projecting object-aware task information onto the environment \citep{andersen2016projecting}. 
These works show that even low-dimensional visual feedback that is grounded in the environment may result in seamless communication. Robot learners can apply lights or projections to indicate when they need more human teaching, to mark their learned behavior, and to indicate how they plan to interact with the environment. 
 
\p{Robot Motion and Gestures} While the previous interfaces add something to the robot or human to act as the visual feedback channel, robots can also use their own motion to convey latent information.
For example, work by \cite{dragan2013legibility} exaggerates the robot's motion towards its goal and away from the other options to more clearly indicate where the robot arm is reaching.
We see this work on legible robot behavior as similar to the implicit communication discussed in Section~\ref{sec:L1}.
The distinction lies in the fact that the robot now intentionally exaggerates or emphasizes its motions to better facilitate communication.
For instance, \cite{kwon2018expressing} propose a framework in which the robot optimizes its behavior to show when it cannot complete a task (e.g., the robot repeatedly starts to lift an object to indicate it is too heavy). The timing of the motion is also useful for communication: \cite{zhou2017expressive} modulate speed to express the robot's internal state along a trajectory.

When the robot has humanoid characteristics, gestures provide another means to visually convey the robot's intent.
Humans are adept at recognizing and interpreting nonverbal cues from robots \citep{venture2019robot, cha2018survey}. 
For instance, robots in industrial settings can give a thumbs up to nearby human workers \citep{sheikholeslami2017cooperative}. 
Within a learning context, \cite{huang2019nonverbal} use the robot's gaze direction to communicate the preferences the robot has learned from a human teacher.
In social robotics, researchers have similarly investigated methods to generate emotional movements or gestures that transmit behavioral intentions \citep{matsumaru2022methods, rossi2021evaluation}.
While gestures and other non-verbal communication carry significant social information, they can be difficult to apply to non-humanoid systems, and may not be sufficient to convey complex concepts. 

\p{Summary} Works on AR, VR, lights, projections, motions, and gestures offer an array of visual interfaces that extend beyond screens and into immersive systems.
Each of these methods can embed visual feedback into the robot's environment (e.g., AR rendering the robot's next waypoint, or a projection showing that the robot is stuck).
User studies and experiments from the surveyed works show that embedding feedback into the environment is more resilient to human motion and changing human perspectives, providing a more natural and interpretable signal for the human.
Additionally, when the feedback is co-located with the task, the human does not need to look in different directions (e.g., turning their head to see a computer screen).

\subsubsection{Towards Non-Visual Feedback Modalities} \label{sec:I1-2}
Immersive visual interfaces render visual data on the spatial context of robot learning scenarios. But for non-visual information, such as forces, and for information that is not spatially embedded, such as alerts, \textit{non-visual interfaces} provide additional benefits \citep{kassem2022happening}. 
Here we focus on \textit{auditory} and \textit{haptic} feedback interfaces for communicating robot learning.
Consider our chair assembly example from Figure~\ref{fig:front}: an auditory sound could alert users when the robot is confused. Alternatively, a haptic band wrapped around the human's wrist could squeeze to indicate how much force the robot has learned to apply when inserting the chair legs.

\p{Auditory Feedback} The primary benefit of auditory interfaces for communicating robot learning is their ability to provide feedback that users can sense from many vantage points.
Humans do not need to be looking in a specific direction or wearing additional equipment to perceive an auditory signal. Additionally, auditory signals can be intuitive for users to interpret, especially if natural language is used. We will separate auditory signals into two categories: verbal, which uses human language to embed meaning in the auditory signal, and non-verbal, which uses artificial sounds (such a ``beeps'' or ``buzzing'') to convey information.

Verbal interfaces have been used to communicate natural language explanations of robot policies \citep{tellex2020robots, hayes2017improving}.
Instead of interacting with a text-based system, humans can ask questions and hear the robot explain its decision-making process.
\cite{schott2023literature} found that verbal explanations --- whether delivered before or after the robot executes an action --- enable the robot to communicate its intent with greater transparency. 
However, the choice of words significantly impacts efficacy. For instance, \cite{struckmeier2019autonomous} show that short, focused verbal communication aids participants in more accurately identifying errors in the robot's policy when compared to more extensive explanations. 
Unchecked use of communication overloads the human’s attention; \cite{unhelkar2020decision} provide a framework for optimally choosing when to speak to the human.
Similarly, the tone of verbal feedback can intentionally or unintentionally communicate emotion (e.g., causing the human to think the robot is upset), and humans often reinforce their speech with non-verbal gestures \citep{prado2012visuo}.
Hence, there are subtle differences between providing verbal feedback or using written language, and simply ``speaking'' a natural language explanation may not always convey the same meaning as a written version of that same text.

Non-verbal auditory feedback has many of the same benefits as verbal feedback, such as transmitting the message regardless of where the human is looking, but it does not require robot learning to be translated into natural language expressions \cite{cha2018effects}. On/off alerts can offer a binary feedback cue \citep{pourmehr2013robust}, or auditory feedback can modulate in tone to provide one-dimensional feedback along a continuous spectrum.
One important aspect of non-verbal interfaces is the way they replicate and rely on social expectations for human-like communication.
For example, \cite{pourmehr2013robust} use societal conventions to generate complex non-verbal sounds that participants naturally interpreted to mean that the robot needs help.
If the robot's sounds are aligned with human conventions, they can rapidly communicate with the user.
Indeed, some work has shown that well designed non-verbal cues outpace verbal communication in helping humans understand a robot's internal state \citep{yamada2006designing}.

\p{Haptic Feedback} Touch encompasses a high-dimensional input for humans, combining often overlapping aspects of pressure, texture, force, and compliance \citep{culbertson2018haptics, pacchierotti2017wearable}. Recent works on communicating robot learning have started to delve into the use of haptics, ranging from human-worn (i.e., wearable) devices to robot-centered (i.e., touchable) interfaces.
\textit{Wearable} haptic interfaces can provide similar benefits to auditory and AR interfaces, and do not require the human's constant attention because the haptic signal is always in contact with the person \citep{battaglia2017rice, baez2023communication}.
For example, in \cite{mullen2021communicating} participants wore a haptic wristband while teaching the robot how to perform a task. Each time the robot was unsure about its next action, the haptic wristband vibrated to cue the human teacher and prompt them to provide expert inputs.
\textit{Touchable} haptic interfaces have the advantage of localizing feedback at a point of interaction, and can therefore associate some spatial meaning with the signal.
For instance, in \cite{valdivia2023wrapping} the researchers wrapped haptic displays around multiple joints of the robot arm.
When humans physically interacted with the robot arm to teach it a task, they naturally touched these different displays along the robot.
If the robot was confused about how to move a specific joint, the robot inflated the haptic displays at that specific joint, conveying both overall uncertainty and spatial location.

Haptic interfaces can be used to communicate multiple types of information, including intent, alerts, and forces.
\cite{che2020efficient} created a hand-held haptic device to communicate a mobile robot's intended path (e.g., what direction the robot planned to move), and \cite{cutlip2021effects, baez2023communication} studied the benefits of haptic feedback for facilitating transfer of driving control between humans and autonomous vehicles. 
Similarly, \cite{casalino2018operator} show that a vibrotactile haptic device can convey when the robot has understood the user's command and intends to follow it, speeding up the completion of collaborative tasks.
Moving beyond alerts, \cite{salvato2021data} demonstrate that wearable haptic devices can also be programmed to convey social touch, with each touch type having a distinguishable meaning to the wearer.
Finally, out of the all interfaces we have surveyed for communicating robot learning, haptic interfaces have had the best success at displaying forces \citep{huang2021human, gong2023haptic, khurshid2016effects}.
For example, \cite{peternel2013learning} developed a full body haptic interface that users wear while teaching a teleoperated robot arm.
This haptic interface replicates and renders the forces experienced by the remote arm, enabling the human teacher to sense the robot's physical state throughout the task.
The human can then use this feedback to adjust their demonstrations; e.g., not pushing too hard on the chair legs when inserting them into the base.

\p{Summary} Overall, the move from screens to alternative visual and non-visual interfaces is motivated by two features:
i) the need to render information in a spatial context, and 
ii) the need to communicate information when the user is distracted or otherwise focused. 
These immersive interfaces also bring new capabilities that are suited for communicating different aspects of robot learning.
This includes verbal feedback to explain robot decisions, projections to convey the robot's intent, or haptic notifications to alert the user.
The surveyed works suggest that there are some representations of robot learning that align well with visual interfaces (e.g., showing the robot's learned trajectory), and there are other representations that align well with non-visual interfaces (e.g., notifying the human when the robot is uncertain).

\subsection{Moving From Single to Multi-Modality} \label{sec:I2}

So far we have surveyed how robots can leverage immersive and non-visual modalities to communicate their learning. 
However, given the diverse representations of robot learning and amount of information the robot learner needs to convey, relying on just a single type of communication interface may not be sufficient. 
Different interface modalities (e.g., visual, auditory, haptic) are best able to convey different types of information. 
Decades of research in the field of human factors have outlined the information transfer capabilities of each human sensor modality, and human factors experts highlight the importance of not overstimulating any one sensory channel \citep{sarter2006multimodal,mortimer2017information, kaber2006investigation}. 
This suggests the promise of multi-modal feedback; to harness these benefits, researchers have developed several different multi-modal interfaces for conveying the robot's latent state and --- more specifically --- for communicating robot learning.
Multi-modal interfaces combine multiple types of feedback working in concert to optimize information transfer back to the human. 

Recent research explores how to distribute the information the robot wants to convey among multiple feedback modes.
\cite{perrin2008comparative} and \cite{sanders2014influence} find that multi-modal interfaces should separate signals based on the timescale (e.g., intermittent vs. real time) and the data type (e.g., discrete vs. continuous).
Consider the example in Figure~\ref{fig:interfaces}. 
As the human teaches the robot, the robot learns both a trajectory to follow and its confidence regarding the waypoints along that trajectory.
The real time and continuous signal (i.e., the planned trajectory) aligns with the capabilities of visual feedback, while the intermittent and discrete signal (i.e., the uncertainty over each waypoint) is suited to the strengths of auditory or haptic feedback.
\cite{mullen2021communicating} implement a similar division between augmented reality and haptics when a human is teaching a robot arm.
Here AR passively visualizes what the robot has learned, while attention-grabbing haptic wristbands actively prompt and direct human teaching. The combination of both augmented reality and haptics lead to better team performance than either AR or haptics alone. 
Looking specifically at communicating robot learning, other works such as \cite{hayes2017improving}, \cite{edmonds2019tale} and \cite{mota2021integrated} use visual feedback in the form of robot motion demonstrations and verbal feedback in the form of explanations to enhance the transparency of robot behavior.

Multiple works have supported the benefits of multi-modal feedback for communicating the robot's latent state.
\cite{bolano2018transparent} show that a mix of visual and auditory feedback allows for a more intuitive interface, immediate understanding of robot actions, and less unpredictability.
Similarly, \cite{bolano2021deploying} demonstrate that humans trust robot partners more when those robots leverage AR and verbal speech. 
As a result, multi-modal feedback is often subjectively preferred by users \cite{han2021need}. 
However, just adding more feedback channels does not guarantee a better interaction --- if not correctly harnessed, multi-modal feedback can become confusing and even negatively impact the user's trust \citep{diethelm2021effects}. 
In what follows will review \textit{three paradigms} for successfully implementing multi-modal feedback in human-robot interaction.

The first successful paradigm leverages different interface modalities for conveying implicit feedback or explicit signals.
This paradigm builds upon the implicit and explicit methods for communicating robot learning discussed in Section~\ref{sec:learning}.
For example, \cite{che2020efficient} and \cite{hagenow2021corrective} mix implicit feedback (robot motion) and explicit signals (audio signals or haptic interfaces) to enhance the robot’s transparency and efficiency. 
\cite{han2021need} and \cite{mirnig2012feedback} apply non-verbal cues such as arm movement, head shake, eye gaze, and facial expression to transfer implicit information related to robot’s behavior or intent. Concurrently, they use verbal explanations to explicitly communicate the robot's decisions and policy.
In each of these studies participants preferred the combination of both feedback types over only implicit feedback.

Another effective paradigm for designing multi-modal communication interfaces focuses on ways to simultaneously convey two distinct information streams without increasing the user's mental workload.
\cite{chu2022multisensory} use visual and haptic feedback, where visual feedback conveys the robot's actions and haptic feedback transmits the robot's contact forces. Along the same lines, \cite{yoon2017customizing}, \cite{pacchierotti2015cutaneous}, and \cite{khurshid2016effects} implement complementary haptic feedback during teleoperation tasks. In \cite{khurshid2016effects} visual feedback was used to highlight important details of the environment, and tactile feedback was utilized to provide guiding forces to the human operator.
Qualitative measures show that users are not subjected to significantly greater mental or physical fatigue when receiving multiple channels, in part because the interfaces are designed to convey completely distinct information with each different modality.

One final paradigm for designing multi-modal interfaces is to separate the robot's feedback into a primary modality (which provides the most important signals) and a secondary modality (which acts in support of the primary modality).
For instance, the secondary modality can provide feedback that reinforces or expands on the information the primary modality is trying to convey.
\cite{diethelm2021effects}, \cite{bolano2021deploying}, and \cite{shrestha2016exploring} use a main channel to communicate the intended robot motion (through speech, AR, or robot motion) supplemented by additional feedback in the secondary channel (gaze, speech, or other auditory cues).
\cite{kassem2022happening} created an interface for environment navigation that communicates the exact same information through three distinct modalities: VR, auditory and haptics. When taken individually, visual feedback performed the best; but the combination of all three modalities outperformed just visual feedback alone.

\p{Summary} Multi-modal communication interfaces --- when designed correctly --- increase the clarity of information transfer, diversify the information stream, and convey more dense data without overloading any one sensory channel. These benefits have been applied to communicate robot learning through combinations of visual, auditory, and haptic signals.
There are several effective paradigms for dividing the robot's information into each separate modality.
Recent works have used multiple interfaces to i) convey implicit and explicit signals, ii) limit the human's mental workload, and iii) reinforce information through multiple channels.
Overall, research on communication interfaces provides an array of effective devices and procedures for seamlessly conveying information to a human.
But just because the human receives the robot's feedback does not mean the human correctly interprets that feedback to build an understanding of the robot learner.
Accordingly, in the next section we survey the the measurement tools used to assess how the human responds to the robot's communication, and how this communication affects the human-robot team.
\begin{figure}[t!]
\centering
\includegraphics[width=1.0\columnwidth]{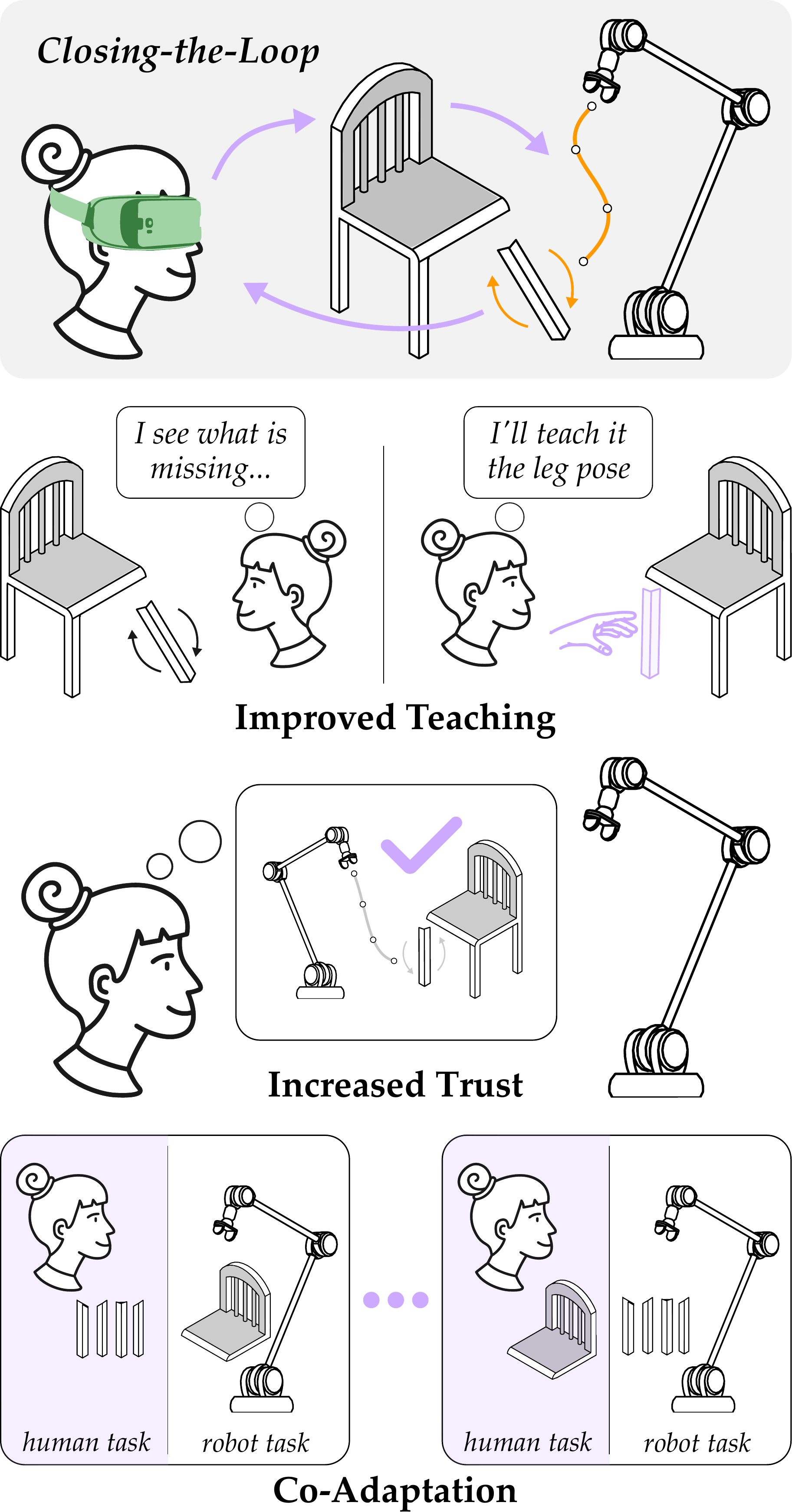}
\caption{Outcomes of closing the loop and communicating the robot's learning to the human teacher. (Top) Human teacher gets feedback that the robot learner is unsure about how to orient the chair leg. (Improved Teaching) Based on the robot's feedback, the human improves their teaching to focus specifically on what the robot is missing. (Increased Trust) This feedback also helps the human align their trust with the robot's learned capabilities. (Co-Adaptation) In multi-agent tasks where the human and robot are working together to assemble the chair, the robot learns from the human's demonstrations, and the human updates their understating of the robot based on the robot's feedback. This can lead to changing roles.} \label{fig:effects}
\end{figure}

\section{Effects of Closing-the-Loop \newline on Robot Learning} \label{sec:models}

In Section~\ref{sec:learning} we surveyed algorithms that intentionally learn in ways that can be communicated, and in Section~\ref{sec:interfaces} we reviewed interfaces that convey that learning back to the human.
Viewed together, learning and communication seek to close the loop so that the human teacher understands what their robot has learned (see Figure~\ref{fig:effects}).
We now return to the human's perspective, and focus on how the human responds to this closed-loop system.
We explore two related research trends:
i) how systems \textit{measure} if the robot's feedback successfully conveys the robot's learning, and 
ii) how closing the loop \textit{impacts} human-robot interaction.

We start in Section~\ref{sec:H1} with state-of-the-art approaches for measuring both the interaction performance and the human's perception of and response to the robot learner.
This includes measures of the human's situational awareness, subjective response, and objective performance.
Related works use one or more of these tools to indirectly probe the human's mental model of the robot; e.g., what the human thinks the robot has learned and how the human predicts the robot will behave.
When robots close the loop --- and provide feedback about their learning --- the human teacher is able to form a more accurate mental model of their robot partner.

Next, in Section~\ref{sec:H2} we survey some common outcomes of closing the loop on robot learning.
These outcomes often stem from the human's improved mental model of the robot learner.
For instance, as the human gets a better sense of what the robot does and does not know, they can focus their teaching on regions where the robot is unsure. 
Experimental results also suggest that --- because humans better understand the robot --- they are more willing to trust the robot when appropriate.
Finally, as the robot learns from the human and the human models the robot learner, both agents can co-adapt to one another.
This co-adaption has led to changing roles and emergent behaviors.

\subsection{Measures of Human-Robot Interaction}\label{sec:H1}
\subsubsection{Interaction Performance}

To measure how communicating robot learning affects human-robot interaction, researchers often consider the performance of the closed-loop system \citep{coronado2022evaluating}.
What is meant by ``performance'' may be task and method specific.
For instance, when a factory worker is teaching their robot arm to assemble a part, a high performance robot may minimize the human's interaction time. 
By contrast, high performance for an assistive robot could correlate to increased usage and longer interactions between the human and robot.
Despite these differences, we have identified some common measurements often used across the literature.
We break these performance metrics down into two categories: measures focused on the \textit{human} and measures focused on the \textit{robot}.

Human-centered metrics seek to quantify when and how the human interacts with the robot.
Common metrics include i) the amount of time the human teaches the robot, ii) the number of inputs the human provides to the robot learner, and iii) the effort associated with the human's inputs (e.g., a minor correction vs. an entirely new demonstration) \citep{sena2020quantifying, pearce2018optimizing}.
If the human and robot are collaborating during the learning process, additional measures of team fluency could include the human's idle time or the delay between when the human expects to interact with the robot and when the robot is ready for interaction \citep{hoffman2019evaluating}.
Questionnaires such as Likert scale surveys \citep{schrum2023concerning} are also relevant here: questions may ask the human about their teaching experience and their perception of the robot's performance. 

Robot-centered metrics assess the performance of the robot learner.
If we let $\theta$ be the parameters that the robot has learned, and $\theta^*$ be the desired parameters the robot should have learned, one straightforward assessment of learning accuracy is the error between actual and desired: $\| \theta^* - \theta\|^2$.
However, just because the model weights are close to the desired weights does not mean the robot will perform the task correctly.
Hence, robot learning often measures \textit{regret}, which captures the difference between the best possible robot behavior and the robot's learned behavior \citep{osa2018algorithmic}.
Regret compares the reward for the optimal trajectory under $\theta^*$ and the reward for the optimal trajectory under $\theta$, where a higher regret indicates more suboptimal behavior.
Outside of error and regret, another common metric is task success.
This metric could be binary (e.g., did the robot assemble the chair correctly?) or measured along a spectrum (e.g., how many legs did the robot autonomously add to the chair?).
When algorithmic efficiency is relevant, metrics may also include the amount of training time, the necessary computational resources, or the number of training iterations.

\subsubsection{Measures of Human's Mental Model} 
As the human receives information through the communication interface, they process this data to form a model of the robot.
We refer to this as the human's \textit{mental model}: this model includes the human's estimate of what the robot has learned and the human's expectations for how the robot will behave when deployed. 
Directly measuring the mental model requires probing human thought, which is difficult if not impossible. Instead, current works aim to indirectly measure the human's understanding of the robot learner by defining standardized and measurable features of the human response \citep{hu2020interact}. Often in scenarios where the robot is conveying its learning these features are measured through offline post-experiment surveys and not online physiological measures.
Accordingly, in this subsection we will first discuss the offline measurement tools commonly used in robot learning research, and then introduce how those measurements are being expanded into reliable online metrics. 

\p{Situational Awareness} One way to assess the human's mental model is the human's awareness of the robot's learned behaviors. This is an instance of situational awareness, i.e., ``a measure of an individual's knowledge and understanding of the current and expected future states of a situation'' \citep{moore2010development}. Situational awareness combines a broad set of features, from attention to understanding to synthesis, and when properly applied these measures can give insight into the user's model of the robot. Methods for measuring situational awareness are classified as direct or indirect \citep{endsley2021situation}, with direct measures, generally self-reported by users through questionnaires, representing the standard measurement approach. In many human-robot learning experiments users will be asked to quantify their agreement with statements that probe their awareness of specific parts of the learning task. For example, the statement ``I think the robot needs my help''~\citep{watkins2021explaining} or ``I could tell what the robot had learned''~\citep{mullen2021communicating} can give a sense of how the human's perception changes as the robot provides feedback.

While these surveys are often task or experiment specific, researchers within the field of human factors have developed standardized survey techniques for direct measurement of situational awareness, such as the Situational Awareness Global Assessment Technique \citep{endsley2000situation}, the Situational Awareness Rating Technique \citep{taylor2017situational}, and the Situation Presence Assessment Method \citep{durso2004spam}. These methods differ in when they query the user (i.e., during or after the task) and whether they query awareness of task details or the user's perception of their own awareness. In a human-robot interaction task involving teleoperation, \cite{schuster2012individual} find that the Situation Presence Assessment Method correlated more with spatial and visual attention than with task performance, while the Situational Awareness Rating Technique correlated best with teleoperation performance. Overall, these survey-based tools offer an established method to compare the effects of different learning algorithms and communication interfaces.
But completely relying on offline surveys can be inefficient within rapidly changing closed-loop systems. 
Hence, physiological real-time measurements of situational awareness may be a future measurement tool for robot learning applications, as we will discuss in Section~\ref{sec:problems}.

\p{Trust} The user's perception of the robot also shapes their trust in that intelligent system. Trust is a step past merely understanding the communication, and explores how humans develop confidence in the reliability of a robot's behavior \citep{kok2020trust}. Like situational awareness, many of the works related to communicating robot learning measure the effects on trust after the interaction (e.g., using survey questions), or they indirectly measure trust from the human's behavior in specific situations (e.g., whether the human follows the robot's suggested trajectory) \citep{chen2020trust, xu2015optimo, pippin2014trust}. Recent research efforts have begun to look at psychophysiological measures as a way to measure trust online and in general applications, including neural signals and physiological reactions. Simple physiological human features --- such as facial expressions and voice analysis --- have been shown to reliably correlate with trust \citep{khalid2016exploring}. These psychophysiological measures are often used in combination with neurological measures to paint a better picture of the human's trust in robots. For example, \cite{hu2016real} and \cite{akash2018classification} combine Galvanic Skin Response and an electroencephalogram to monitor trust levels in real time and develop trust sensor models for intelligent machines that build and maintain human trust during interactions.

\p{Mental Workload} The user's perception of the robot only measures part of the feedback's effects; another aspect is the amount of human mental effort required to sense and interpret the robot's signals. 
Ideally, when robots close the loop and communicate their learning, the human will not need much time or thought to parse these signals and infer what the robot knows.
In practice, mental workload is often inversely related to situational awareness.
For example, \cite{dini2017measurement} show that more information-dense feedback can improve the human's situational awareness, but at the expense of increased mental workload when processing the dense feedback. Offline measures of the human's workload once again take the form of surveys, with the NASA-TLX \cite{hart1988development} forming a standard approach to measuring different features of physical and mental workload \citep{memar2019objective}. 
Indirect measures like heart rate have also been shown to be good assessments of mental workload \citep{mach2022assessing}, but may require significantly higher levels of human effort than typically exist during human-in-the-loop robot learning to identify measurable effects. For robot learning, online measures with clear correlation and large effect size are more applicable to measure workload; these include eye-tracking \citep{devlin2022scan} and electroencephalograms \citep{novak2015workload, memar2019objective, hogervorst2014combining}.

\p{Summary} We can measure the effects of closing the loop and communicating robot learning by looking at interaction performance and probing the human's mental model.
Within interaction performance, common metrics include the error and regret in the robot's learned behavior.
To measure the human's perception of the robot learner, researchers often account for situational awareness, trust, and workload.
Each of these aspects can be quantified offline through the standardized use of survey techniques. There is a recent thrust towards online measurements tools (such as tracking gaze or heart rate), but these methods have not been widely incorporated within current robot learning systems.

\subsection{Outcomes for Human-Robot Interaction} \label{sec:H2}

Once we close the loop by communicating the robot's learning back to the human teacher, how does this change the team's behavior and the human's experience?
The measures outlined in Section~\ref{sec:H1} provide an array of tools to estimate the subjective and objective effects of communicating robot learning.
In what follows, we survey some of the common outcomes these measurements have identified.
We find three research themes across recent studies: these works suggest that closing the loop can improve human teaching, increase human trust, and facilitate co-adaptation (see Figure~\ref{fig:effects}).

\p{Improved Teaching}
Communicating the robot's learning reveals what the robot has and has not learned.
As the human teacher processes this feedback, they can adjust their teaching to focus specifically on the parts of the task where the robot is still confused.
For instance, consider a human teaching an autonomous car how to drive on highways.
Without any feedback, the human might provide unfocused, random demonstrations that show how to pass other cars, change lanes, and merge on and off the road.
But after the robot implicitly or explicitly communicates that it is most uncertain about merging, the human can now focus on giving multiple examples of merging trajectories \citep{spencer2022expert, tian2023towards}.
In this way, communication can help focus teaching directly on what the robot is trying to learn.

From the human's perspective, there are multiple axes along which users can adjust their teaching.
These include when to provide guidance, what type(s) of inputs to provide, and which parts of the task to teach.
Different methods for communicating the robot's learning align with different teaching axes. 
For example, saliency methods such as \cite{watkins2021explaining} and \cite{olson2021counterfactual} use visual interfaces to highlight \textit{where} the human should provide additional demonstrations; by contrast, 
the haptic wristband used by \cite{mullen2021communicating} notifies the human \textit{when} the robot is confused and needs guidance.
There are benefits to each axes.
Recent experiments by \cite{sena2020quantifying} indicate that communicating when, where, and how to provide demonstrations reduces the human's mental burden and leads to more focused human teaching.

From the robot's perspective, focusing the human's inputs on areas of uncertainty can accelerate robot learning (i.e., the robot can learn the task from fewer demonstrations).
Approaches from Section~\ref{sec:learning} such as human-in-the-loop reinforcement learning \citep{lee2021pebble} and active preference-based learning \citep{sadighactive} display behaviors or trajectories where the robot is unsure, and ask for the human's inputs specifically in these regions.
When compared to baselines in which the human selects their preference from a set of randomly sampled trajectories, proactive methods infer the human's task more accurately and efficiently \citep{lee2021pebble, sadighactive, tucker2020preference}.
\cite{biyik2022learning} provide theoretical support for these results: 
to optimize learning the robot should first collect any unstructured, open-ended human demonstrations, and then elicit specific human feedback about the remaining areas of uncertainty.
Overall, when the human's inputs focus on what the robot does not know, existing learning frameworks more accurately infer the desired task.

\p{Increased Trust}
A second potential benefit of closing-the-loop on robot learning is increased user trust in the system. 
Trust can be described as a psychological condition in which the person's inclination is to rely on a robot to complete the task \citep{madsen2000measuring}. 
In human-robot interaction, prior works have explored how robots build and maintain trust with humans \citep{khavas2020modeling}, and how robots can rebuild trust over time after that trust is violated \citep{de2020towards, baker2018toward}. 
\cite{hancock2011meta} identify several factors that affect trust: i) human-related (e.g., ability), ii) robot-related (e.g., performance), iii) environmental (e.g., team collaboration), and iv) task-related (e.g., type or complexity). Communicating robot learning falls within the \textit{robot-related} category. 
More specifically, the robot's feedback helps the human anticipate how the robot will perform, where it will succeed, and when it might fail.
This increased transparency and predictability may promote the human's trust in the robot learner.

However, as we described in Section~\ref{sec:H1}, trust remains difficult to measure or quantify.
Works such as \cite{freedy2007measurement} and \cite{gao2013modeling} formulate trust based on performance; i.e., trust is correlated with the outcomes of human-robot collaboration, and the more frequently the human intervenes the less they trust the robot \citep{xu2016towards}.
Other approaches treat the human's trust as a latent variable, and infer that variable from the human's actions throughout the task \citep{chen2020trust, xu2015optimo}.
For example, if the human delegates a more difficult role to the robot learner (i.e., asking the robot to carry a glass jar), the human may trust the robot more \citep{pippin2014trust}.
With these different measures in mind, we do not claim that communicating robot learning always increases one specific definition of trust.
Instead, we find trends across related studies to suggest that communication generally enhances trust or trust-adjacent metrics.
Studies in \cite{zhu2020effects} show that intelligent robots who proactively explain their decision-making before taking actions can build trust with humans. 
Similarly, displaying a confidence signal \citep{desai2013impact}, rendering real-time information on a GUI \citep{boyce2015effects}, providing multi-modal feedback \citep{ciocirlan2019human} and using mixed reality interfaces \citep{rosen2020mixed} have all been shown to increase different measures of human trust in an intelligent robot.
Overall, these experimental findings indicate that robots which reveal their learning to humans will better align the human's expectations with the robot's capabilities, and increase the human's trust in how the system will behave.

\p{Human-Robot Co-Adaptation}
One final outcome that has been measured in systems that communicate robot learning back to the human teacher is co-adaptation.
This especially applies to collaborative or competitive robots that learn while interacting with the human in multi-agent tasks.
Consider an extension of our example from Figure~\ref{fig:front} where the human and robot are now working together to assemble chairs.
Perhaps for the first few chairs the robot holds the base, and the human demonstrates how to add legs to that base (see Figure~\ref{fig:effects}).
But as the robot learns from the human --- i.e., as the robot learns how to add the chair legs --- the robot can adapt its behavior to better collaborate with the human.
Similarly, as the human observes the robot's implicit and explicit feedback and understands the robot's capabilities, the human can also co-adapt alongside the robot \citep{mortl2012role, van2021becoming}.
Over time the human and robot might switch roles, so that the human holds the base and the robot adds the legs.
Co-adaptation occurs here because the robot learns from the human, the human gets feedback about what the robot has learned, and both agents adjust to more seamlessly complete the interactive task.

Related works study how robots can communicate their learning --- or take other types of actions --- to encourage co-adaptation during multi-agent tasks.
These approaches often frame co-adaptation as an optimization problem \citep{nikolaidis2017human, pellegrinelli2016human, parekh2022learning}.
Under this framework the robot learns a predictive model of how the human will behave during the task.
The robot then selects its own feedback signals or actions so that, when these actions are paired with the learned human model, the robot optimizes its cumulative reward.
For example, in \cite{nikolaidis2017human} the robot models the human's willingness to adapt as a latent parameter, and solves a partially observable Markov decision process to find the policy that both infers this latent parameter and coordinates with the human.
If the human is not willing to adapt the robot follows the human's lead; but if the human does adapt, the robot intentionally causes the human to adapt in such a way that the human-robot team completes the task more quickly and efficiently.
\cite{xie2021learning} and \cite{parekh2022learning} extend this approach to construct more general latent representations of the human.
Within \cite{parekh2022learning} the robot plays tag with human users: as the robot learns where the human will ``hide,'' the participants also get a better sense of where the robot will ``seek.''
This results in co-adaptive behaviors where the learning robot continually updates its model of the human, and the human reactively changes their behaviors to avoid the robot opponent.

Co-adaptation can also occur at the communication level.
Different users respond to the same feedback signals in different ways, and as the robot learns from the human it should adapt its signals to better coordinate with the user.
In \cite{zhao2022role} the robot recognizes that the human may not correctly understand every communicated signal, and so the robot modifies its feedback to account for the human's interpretation.
\cite{chen2022mirror} learn a model of the human operator, and then use that model to determine what, when, and how to communicate with that user by simulating the human's potential future actions.
Similarly, both  \cite{reddy2022first} and \cite{christie2023limit} seek to adapt the robot's communication interface to align the robot's signals with the human's response.
Here the robot does not know exactly how the human will interpret its haptic, visual, or audio cues; hence, the robot continually adjusts the way it selects these cues based on its learned model of the human.
From the user's perspective, as the human gets experience working with the robot they better understand what each robot signals means.
From the robot's perspective, as the robot sees the human react to its signals, it can adjust its future signals to more accurately convey its learning.

\p{Summary}
In this section we surveyed how closing the loop with learning and communication can affect the human and overall system.
Measurement tools and user study results suggest that i) we can monitor the human's mental model of the robot learner and ii) communicating robot learning significantly impacts human-robot interaction.
Benefits of conveying the robot's learning include improved human teaching, aligning trust with the system's capabilities, and co-adaptation to more seamlessly complete collaborative tasks.
To measure these benefits --- and probe the human's underlying interpretation of the robot's signals --- we turn to measures of situational awareness, post-hoc surveys, and objective performance metrics.
\section{Open Questions and Future Directions} \label{sec:problems}

In Sections~\ref{sec:learning}--\ref{sec:models} we reviewed trends in human-robot interaction at the intersection of learning and communication.
When viewed together, this body of research offers a set of tools to close the loop, communicate robot learning to human teachers, and assess the outcomes of this feedback.
However, there remains a significant knowledge gap between what the robot has learned and what the human thinks the robot has learned.
State-of-the-art robot learners can be confusing or misleading, even when they follow the surveyed methods \citep{habibian2022here, kessler2021interactive}.
The ongoing and diverse research in this area suggests that there remain open questions that must be addressed before we reach human-robot systems where the human completely understands what their robot is learning.

\begin{figure}[t!]
\centering
\includegraphics[width=1\columnwidth]{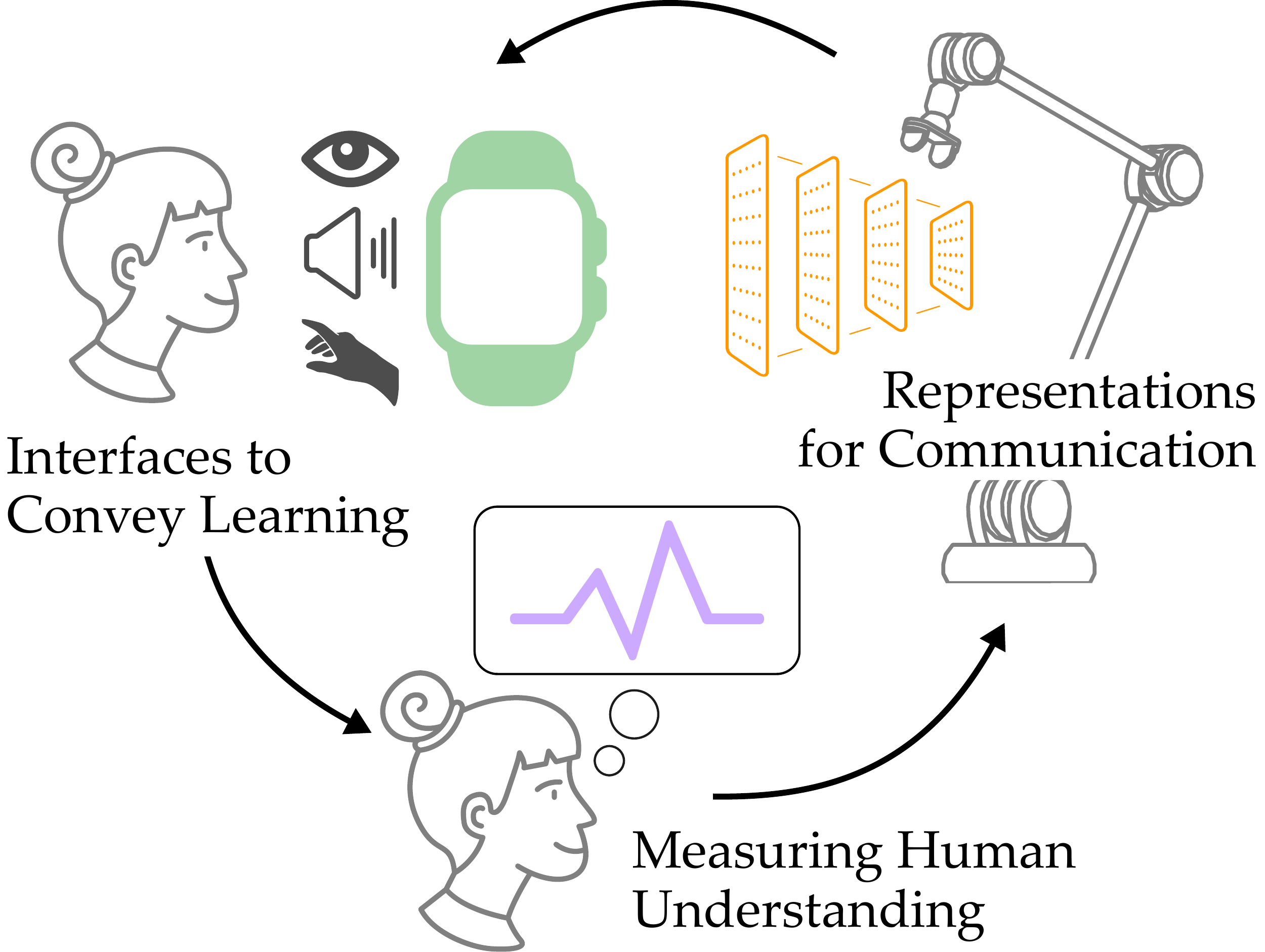}
\caption{Open questions for communicating robot learning. To advance the state-of-the-art, we seek (Section~\ref{sec:p1}) robots that design their learning representations to align with mutli-modal communication interfaces, (Section~\ref{sec:p2}) standardized communication interfaces that are capable of conveying robot learning, and (Section~\ref{sec:p3}) online measurement and modeling tools to estimate how the human interprets the robot's feedback. 
} \label{fig:problems}
\end{figure}

In this section we propose and discuss a set of open questions for communicating robot learning (see Figure~\ref{fig:problems}).
We emphasize that these challenges are interdisciplinary, and will not be solved by just improving the learning algorithm or communication interface in isolation. 
Instead, solutions likely lie at the intersection: 
research on robot learning that is aware of the capabilities of communication interfaces,
and research on interfaces that understands the types of data robot learners need to convey.
For example, a ubiquitous challenge is the trade-off between low- and high-dimensional feedback.
On the one hand, low-dimensional feedback provides a simplified representation of the robot's learning that is easy for humans to quickly interpret.
On the other hand, high-dimensional feedback enables the robot to convey a more comprehensive and accurate view of its learned models.
Determining the right balance between low- and high-dimensional representations requires perspectives on learning (e.g., how can we succinctly and intuitively capture the robot's learning?) and perspectives on communication (e.g., how can we create signals that rapidly and clearly convey complex information to the human?).

%%% Learning problem
\subsection{How Should Robots Convert their Learning into Feedback Signals?} \label{sec:p1}

When robots learn from humans, they often build neural network models that include thousands of parameters.
The learning approaches surveyed in Section~\ref{sec:learning} recognize that robots cannot communicate all of these parameters.
Instead, robots must identify compact \textit{representations} of their high-dimensional learning that can be intuitively conveyed to the human teacher \citep{bobu2023aligning}.
These learning representations should be designed to harness communication interfaces, and provide clear, interpretable, and real-time signals to the human.

\p{Reasoning about Implicit Communication} 
Before the robot selects feedback signals to summarize its learning, we first consider the information the robot \textit{implicitly} communicates during its learning process.
Implicit communication occurs within human-in-the-loop learning frameworks: we have surveyed examples from reinforcement learning, active learning, and imitation learning (see Section~\ref{sec:L1}).
Each of these frameworks follow an iterative process.
During an iteration the human observes the robot's behaviors and provides their expert labels: 
the robot learns from these labels, while the human obtains implicit feedback by watching how the robot's behavior changes over time.

We propose two related questions for advancing implicit feedback.
Many recent works that incorporate a human teacher focus on robot learning, and do not directly consider how the human teacher might interpret the robot's behaviors during the learning process \citep{lee2021pebble, kelly2019hg, losey2022physical}.
This leaves implicit communication as an unintended and uncontrolled consequence.
Accordingly, our first question is how can we best account for implicit feedback within interactive learning algorithms?
As we start to answer this question, we recognize a potential trade-off between efficient learning and implicit communication.
Learning methods often seek to reduce the amount of human involvement so that the human does not need to spend as much time teaching the system --- i.e., the human provides fewer labels, demonstrations, or corrections \cite{hejna2023few, hoque2021thriftydagger, spencer2022expert, jin2022learning}.
An advantage of these works is that the robot learns rapidly from less human guidance.
But from a communication perspective, decreasing the human's involvement means less opportunities for the human to implicitly observe what the robot has learned and what the robot is still confused about.
This suggests that future algorithms which account for implicit communication should balance human involvement.
For instance, the robot could intentionally choose behaviors that both gather information from the human and reveal what the robot has learned so far \citep{habibian2022here}.

\p{Designing Representations for Interfaces}
Moving beyond implicit communication, we next consider \textit{explicit} frameworks that intentionally pre- or post-process their learning to extract feedback signals.
By rendering these feedback signals to the human the robot can directly explain its learning (e.g., reveal why it is taking an action or where it is uncertain).
The state-of-the-art methods we reviewed in Section~\ref{sec:L2} extract signals such as a sequence of waypoints, an image of critical states, or a graph of robot decisions.
To actually convey these signals current robots primarily rely on visual interfaces and computer monitors \citep{watkins2021explaining, huang2019enabling, kenny2023towards, liu2018interactive, olson2021counterfactual}.
But separate research on communication has shown the benefits of more immersive and multi-modal interfaces (see Section~\ref{sec:interfaces}).
This leads to our next open question: how can we design explicit learning frameworks to take advantage of diverse interface capabilities?

Answering this question may require multiple perspectives.
Instead of first selecting a way to represent robot learning, and then looking for interfaces that can convey that representation, designers may need to first identify the communication interfaces that are available to the robot, and then identify the learning representations that align with those interface capabilities.
For instance, the dimensionality of the pre- or post-processed learning signal could depend on the interface modality.
In scenarios where the robot is given a graphical user interface to communicate its learning, high-dimensional and continuous signals are possible: e.g., the robot could render visual waypoints to indicate the locations it has learned to reach.
In settings where the communication interface uses haptic or auditory cues, low-dimensional and discrete signals are suitable: e.g., a wearable haptic device could vibrate when the robot enters regions of uncertainty.
But this is just one possible solution.
Future works will need to determine how best to extract learning representations that multi-modal interfaces can clearly convey.

% Interface problem
\subsection{How Should We Design Interfaces to Communicate Robot Learning?} \label{sec:p2}

In the same way that we want the learning representation to align with the communication interface, we also want to develop physical interfaces that can seamlessly communicate robot learning signals.
In Section~\ref{sec:interfaces} we surveyed a variety of existing interfaces.
Examples include augmented reality headsets to display a drone's trajectory \citep{walker2018communicating}, tactile haptic arrays to indicate the direction of a mobile robot \citep{che2020efficient}, or natural language sentences to explain a robot arm's failure modes \citep{tabrez2019explanation}.
Many of the interfaces that we identified were designed to convey specific latent states (e.g., goals, directions, or failure modes).
Moving forward, the community will need multi-modal interfaces that are purposely created to communicate a spectrum of robot learning representations.

\p{Establishing Standardized Interfaces} 
Overall, our vision here is for future work to establish a \textit{standardized} set of interfaces that the community uses to communicate robot learning.
These standardized interfaces could consist of multiple building blocks (e.g., visual, haptic, and audio modules), as well as guidelines for the types of information that each building block best conveys to the human.
Other robotics applications have already established standards for interface design: 
for instance, there are interface standards for semi-autonomous vehicles \citep{human2018factors}.
Creating similar standards here would be particularly helpful for robot learning researchers. 
Knowing what types of information the interfaces can convey --- and the recommended protocol for conveying that information --- will help designers select the right robot learning representations.
In addition, having a standardized set of interfaces will better enable comparisons between the different methods in Section~\ref{sec:L2} for converting the robot's learning into explicit feedback signals.

To achieve this vision there are multiple open questions.
First, we need to ensure that the standardized interfaces are capable of communicating the information that captures robot learning.
Our surveyed works suggest that conveying the robot's learning goes beyond simply indicating the robot's goal or next action \citep{mullen2021communicating, hayes2017improving, valdivia2023wrapping, huang2019enabling, watkins2021explaining, french2019learning}.
Instead, the interface must communicate more abstract and complex concepts such as uncertainty over an action, reasons behind a decision, or features of a policy.
Based on the trends from Section~\ref{sec:interfaces}, we anticipate that effectively communicating data like uncertainty, reasons, or features will require multi-modal visual and non-visual systems.
Additional interface testing, user studies, and psychometric analysis will be needed to determine whether each individual or combined interface modality can convey these signals.
For example, is a haptic notification sufficient to convey the robot's level of uncertainty?
Follow up questions include the timing of the signals (e.g., how often should the robot attempt to convey its learning to the human?) and the dimensionality of the signals (e.g., should the robot provide binary feedback or feedback along a continuous spectrum?).
Answering these questions will likely include perspectives on i) what the robot learner needs to convey, ii) how we can design interfaces to convey that information, and iii) how we perform and analyze user studies to quantify whether the interfaces were successful.

%%% Outcomes problem
\subsection{How Should We Measure the Human's Understanding of Robot Learning?} \label{sec:p3}

One goal of communicating robot learning is to reach mutual understanding between the human teacher and robot learner.
As the robot learns from the human, the robot updates a model of the task the human wants it to complete.
As the robot communicates back to the human, the human forms a model of what the robot will do when it is deployed.
The research reviewed in Section~\ref{sec:models} suggests that we are moving towards this goal: current approaches that close the learning loop have accelerated robot learning, augmented human trust, and increased co-adaptation \citep{sena2020quantifying, mullen2021communicating,  chen2020trust, parekh2022learning, nikolaidis2017human, watkins2021explaining, kenny2023towards}.
However, it is still not clear if and when a human teacher and robot learner reach mutual understanding.
To better assess the outcomes of communicating robot learning, we need \textit{measurement tools} and \textit{human models} that capture how users interpret the robot's feedback signals and form mental models of the robot learner.

\p{Measuring Human Understanding in Real Time}
The human teacher forms a mental model of what the robot has learned based on the robot's implicit and explicit feedback.
For example, by reasoning over augmented reality displays, haptic notifications, or natural language explanations, the human might estimate that the robot arm in Figure~\ref{fig:front} knows how to carry and insert chair legs.
One open question here is how we best \textit{measure} the human's understanding of the robot's learning in order to improve communication.
In Section~\ref{sec:models} we surveyed existing metrics such as situational awareness, trust, and workload \citep{endsley2021situation, chen2020trust, memar2019objective}.
However, there are two issues with these metrics.
First, current measurement approaches often rely on offline data, such as questionnaires administered after the interaction is over.
Second, it is not fully understood how online measurement tools correlate with the underlying mental model: e.g., if the human's eye gaze is aligned with the robot's goal, does that mean the human understands what the robot has learned?

Future research can address these questions by identifying real-time variables that are connected with the human's underlying model of the robot learner.
Recent work in human factors suggests that we can measure the human's understanding of a situation through neurological measures like electroencephalograms (EEG) \citep{akash2018classification, jung2019neural} and functional near-infrared spectroscopy (fNIRS) \citep{goodyear2016advice}.
For example, \cite{kohn2021measurement} provide experimental evidence that EEG and fNRI are a reliable real-time measure of trust in automation-related applications.
Similarly, \cite{heard2018survey} show that a variety of online metrics for measuring the human operator's workload are available, but in their current form they may not generalize to different tasks or users.
Determining which --- if any --- of these measures correlate to the human's mental model during robot learning will help us assess which explicit feedback signals clearly convey the robot's learning, and which signals are less interpretable by the human.

\p{Updating Human Models}
The tools described above have the potential to measure the human's current understanding of the robot learner.
But how will the human's understanding change over time as they receive new feedback signals?
Put another way: if the communication interface renders a given signal, can the robot predict how the human teacher will interpret and react to that signal?
Here we seek \textit{human models} that relate robot communication to the human's understanding of the robot learner.

These human models may build upon ongoing research on machine teaching (see Section~\ref{sec:learning}).
Within machine teaching the robot intentionally selects behaviors or signals to convey information to a human observer \citep{huang2019enabling, lee2021machine, brown2019machine, cakmak2012algorithmic}.
For example, in \cite{huang2019enabling} an autonomous car generates a sequence of videos that --- when watched by the human --- convey how aggressively the autonomous car drives.
Recent works hypothesize that the way the human updates their understanding in response to these videos can be modeled using Bayesian inference \citep{tenenbaum2011grow} or gradient descent \citep{liu2017iterative}.
But communicating robot learning adds additional complexity to this problem: 
because the robot is learning, the information the robot is trying to convey continually changes.
Future works will also need to extend these models beyond visual feedback (e.g., videos) to incorporate the non-visual and multi-modal signals that might be provided by feedback interfaces.
If successful, improved human models could offer a predictive tool that researchers in learning and communication leverage to compare different signals and develop standardized feedback interfaces.

% Overall summary
\p{Summary}
In some ways conveying the robot's learning back to the human teacher is as challenging as learning the human's desired task in the first place.
Humans are inherently variable: signals, modalities, and information densities that convey the robot's learning to one user could be confusing to another user.
As such, there may be no single answer to the open questions we have listed above.
Instead, we encourage researchers to explore interdisciplinary approaches that cross-over between learning representations, communication interfaces, and human measurements.
\begin{figure*}
\centering
\includegraphics[width=2\columnwidth]{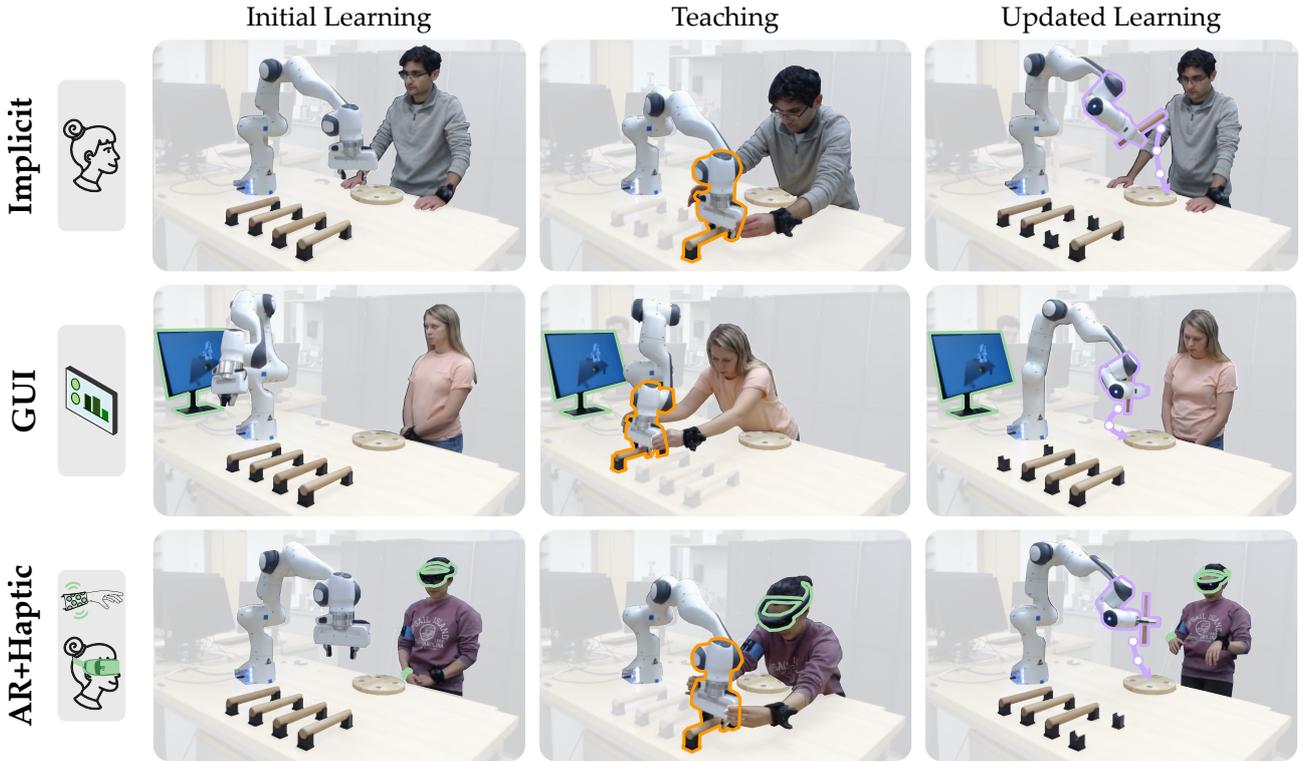}
\caption{Experimental setup for the Case Study in Section~\ref{sec:userstudy}. Participants taught the robot arm to assemble a chair by placing wooden legs into the base. The robot learned using an interactive imitation learning algorithm. In \textbf{Implicit} the robot communicated what it had learned only through its behavior. For \textbf{GUI} and \textbf{AR+Haptic} the robot leveraged a saliency method to extract a representation of its learning; the robot then explicitly conveyed this representation through a GUI or augmented reality and haptic interfaces (highlighted in green). In the first column the robot starts to perform the task given its incomplete initial learning, choosing wrong actions and grasping far from the intended chair legs. In the second column the participant notices the mistake (using either the implicit or explicit feedback) and shows the robot the correct waypoint. Finally, the third column visualizes the robot's resulting behavior after learning from the human's input. Ideally, the robot should place the leg in the spot marked by the purple arrow. Notice that with \textbf{Implicit} the robot has not learned the correct orientation and insertion point for the leg.} \label{fig:user}
\end{figure*}

\section{Case Study} \label{sec:userstudy}

In Sections~\ref{sec:learning}, \ref{sec:interfaces}, and \ref{sec:models} we identified recent research trends in work on communicating robot learning to human operators.
To demonstrate some of these trends in practice, we will conclude our review article with a new case study.
In this case study $12$ in-person participants kinesthetically taught a $7$ degree-of-freedom Franka Emika robot arm how to assemble a simple chair (see Figure~\ref{fig:front} and Figure~\ref{fig:user}).
The robot arm did not initially know where to pick up the chair legs or how to insert them into the chair base.
Users physically intervened to correct the robot's motion, and the robot learned from these corrections to improve its task performance.
We implemented example learning algorithms that provided implicit feedback to the human teacher (Section~\ref{sec:L1}) and converted the learned models into explicit signals (Section~\ref{sec:L2}).
To convey the explicit feedback to the human, we tested interfaces that offered immersive visual displays (Section~\ref{sec:I1}), as well as multi-modal interfaces for communicating the robot's learning (Section~\ref{sec:I2}).
Finally, to assess the outcomes of this case study, we used objective measures of task performance and subjective measures of the human's response (Section~\ref{sec:H1}).
Our measurements were designed to quantify changes in human teaching, trust, and co-adaptation (Section~\ref{sec:H2}). Videos of the case study can be found here: \url{https://youtu.be/EXfQctqFzWs}.
In addition, the code we used to conduct this study is available at: \url{https://github.com/VT-Collab/communicating-robot-learning}

\subsection{Experimental Setup}

Users taught the robot arm under three different conditions (see Figure~\ref{fig:user}).
In the first condition the robot applied a human-in-the-loop learning framework to implicitly convey what it was learning to the human teacher (\textbf{Implicit}).
For the next two conditions the robot structured its learning algorithm to extract explicit feedback signals.
In \textbf{GUI} the robot displayed a visual representation of the key waypoints it had learned on a computer monitor.
In \textbf{AR+Haptic} the robot displayed the same information as in \textbf{GUI}, but now using an augmented reality headset and wearable haptic wristband.
Viewed together, \textbf{Implicit}, \textbf{GUI}, and \textbf{AR+Haptic} are examples of implicitly and explicitly conveying robot learning, and of using visual, immersive, and multi-modal interfaces for communicating the explicit signals.

\p{Learning Algorithms}
The robot arm learned the human's desired policy using interactive imitation learning.
Specifically, we applied human-gated DAgger \citep{kelly2019hg}.
Under this approach the robot's policy $a_r = \pi_{\theta}(s)$ is instantiated as a multi-layer perceptron with weights $\theta$. This network inputs the measured system state $s$ (i.e., the pose of the chair legs and base) and outputs robot actions $a_r$ (i.e., waypoints in joint space for the robot to reach).
Let $\mathcal{D} = \{(s_1, a_1), \ldots, (s_n, a_n)\}$ be a dataset of state-action pairs provided by the human expert.
The robot trains its policy $\pi_{\theta}$ so that the robot's actions $a_r$ match the actions of the human expert $a$.
More formally, the robot learns weights $\theta$ to minimize the loss function:
\begin{equation} \label{eq:C1}
    \mathcal{L}(\theta) = \sum_{(s, a) \in \mathcal{D}} \| \pi_\theta(s) - a \|^2
\end{equation}

Within human-gated DAgger the human can intervene at any time to teach the robot.
For instance, if the participant notices that the robot is reaching for the wrong chair leg, they can stop the robot's motion and show it the correct waypoint at the current state.
The robot adds this new $(s,a)$ pair to dataset $\mathcal{D}$ and retrains its policy to minimize Equation~\ref{eq:C1}.
As the robot collects new data points, its policy should converge to the human's desired behavior.
The robot \textbf{Implicitly} conveys what it is learning through its actions: if the robot makes a mistake, the human teacher can infer that the robot is still uncertain about that part of the task, or that the robot has learned to do that part of the task incorrectly.

\begin{table*}[t]
\caption{Questions on our Likert scale survey. The questions were grouped into six scales: Learned, Trust, Adapt, Intuitive, Easy, and Prefer. On the right we report the reliability of each multi-item scale (Cronbach's $\alpha$) and the results of a repeated measures ANOVA. Here a $p<.05$ indicates that the differences in the users' scores across methods were statistically significant.}
\label{table:likert}
\centering
    \begin{tabular}{lcccc}
        \hline Questionnaire Item & Reliability & $F(2,22)$ & p-value \bigstrut \\ \hline 
        \bigstrut[t]
        -- It felt like I had to repeatedly teach the same thing before the robot understood. & \multirow{2}{*}{$.83$} & \multirow{2}{*}{$6.53$} & \multirow{2}{*}{$<.05$} \\  -- The robot quickly \textbf{learned} what I wanted it to do. \bigstrut[b] \\ \hline  
        \bigstrut[t]
        -- By the end of the interactions, I could \textbf{trust} the robot to do the task correctly. & \multirow{2}{*}{$.94$} & \multirow{2}{*}{$3.55$} & \multirow{2}{*}{$<.05$} \\ -- At the end of the experiment I still did not trust the robot. \bigstrut[b] \\ \hline  
        \bigstrut[t]
        -- I adjusted how I worked with the robot over time. & \multirow{2}{*}{$.64$} & \multirow{2}{*}{$0.37$} & \multirow{2}{*}{$.70$} \\ -- I did not \textbf{adapt} to the robot. \bigstrut[b] \\ \hline  
        \bigstrut[t]
        -- I could tell what the robot was learning and what it was still confused about. & \multirow{2}{*}{$.76$} & \multirow{2}{*}{$14.74$} & \multirow{2}{*}{$<.001$} \\ -- It was not \textbf{intuitive} at all what the robot was learning. \bigstrut[b] \\ \hline  
        \bigstrut[t]
        -- I could \textbf{easily} interpret the robot's feedback and figure out what it wanted to say. & \multirow{2}{*}{$.85$} & \multirow{2}{*}{$16.45$} & \multirow{2}{*}{$<.001$} \\ -- I had to think carefully about the robot's feedback to determine what it meant.\bigstrut[b] \\ \hline  
        \bigstrut[t]
        -- Overall, I \textbf{prefer} this condition & \multirow{1}{*}{$-$} & \multirow{1}{*}{$7.98$} & \multirow{1}{*}{$<.01$} \bigstrut[b] \\ \hline   
    \end{tabular}
\vspace{-0.5em}
\end{table*}

Next, we extended this human-in-the-loop approach to extract explicit signals about the robot's learning (\textbf{GUI} and \textbf{AR+Haptic}).
More specifically, we applied a saliency method from Section~\ref{sec:L2} that highlighted regions of the task where the robot was unsure about the correct action \citep{watkins2021explaining}.
Under this post-hoc framework the robot maintained an ensemble of $N$ models:
\begin{equation} \label{eq:C2}
    \mathcal{E} = \{ \pi_{\theta_1}, \pi_{\theta_2}, \ldots, \pi_{\theta_N} \}
\end{equation}
Each individual model was trained using human-gated DAgger: the models were initialized with randomly sampled weights, and each separate model updated its own weights to minimize the loss function from Equation~\ref{eq:C1}.
The saliency method then post-processed the ensemble of trained models $\mathcal{E}$ to extract feedback signals.
At a given state $s$, the robot queried each of its $N$ models to determine their actions $a_{r,1}, a_{r,2}, \ldots a_{r_N}$. 
If each model agreed on the action (i.e., if the standard deviation over actions was below a threshold $\eta$) then the robot was confident.
By contrast, in states where the models diverged, the standard deviation over actions increased to indicate that the robot was unsure.
In our experiment we set the threshold $\eta = 0.45$.

The overall output of this learning algorithm for explicit communication was i) a prediction of the robot's next waypoint $a_r$ obtained by averaging over the ensemble of actions, and ii) a binary value representing whether or not the robot was confused obtained from the standard deviation over the ensemble of actions. We provided these two explicit feedback signals to the \textbf{GUI} and \textbf{AR+Haptic} communication interfaces in real time.

\p{Communication Interfaces}
To convey signals that represented the robot's learning back to the human participants we tested two different interfaces (in addition to the implicit communication given by the robot's movement).
Each of these explicit feedback interfaces communicated the same information; however, the interfaces were in different locations and leveraged different modalities.

Under \textbf{GUI} we displayed representations of the robot's learning on a computer monitor placed next to the participant and robot arm (see Figure~\ref{fig:user} and our \href{https://youtu.be/EXfQctqFzWs}{video}).
This computer interface showed a physics rendering of the Franka Emika robot.
The rendering was updated in real time to align the simulated robot's state with the state of the real system.
To convey the explicit signals extracted from the learning algorithm, the interface used colored spheres.
The \textit{location} of these spheres changed during the task to mark the robot's learned waypoints (i.e., where the robot had learned to reach).
The \textit{color} of these spheres also changed to reflect the robot's confidence: when the robot was confident about a waypoint, the sphere was green, and when the robot was uncertain about a waypoint, the sphere was red.
By looking at the \textbf{GUI}, participants could see both i) the waypoints the robot had learned along the task and ii) how certain the robot was about each waypoint. Overall, \textbf{GUI} represented a conventional screen-based interface, typical in human-robot interaction (Section~\ref{sec:I1-1}). This type of interface offered participants a familiar and accessible channel for receiving the robot's feedback. 

With \textbf{AR+Haptic} we conveyed the same information as in \textbf{GUI}, but we communicated that information in a different way.
\textbf{AR+Haptic} represented a shift from the computer monitor to more immersive and accurate projections in the task space (Section~\ref{sec:I1}), while also exploring the benefits of multi-modal feedback (Section~\ref{sec:I2}).
Similar to \cite{mullen2021communicating}, this immersive and multi-modal approach combined both a wearable augmented reality headset (Microsoft HoloLens $1$) and a wearable haptic wristband.
The wristband was developed by the authors based on our prior work \citep{valdivia2023wrapping}.
The band was composed of a row of thin-walled inflatable plastic pouches: increasing the pneumatic pressure caused the pouches to inflate and apply normal forces to the participant's wrist.
We used this haptic wristband to notify the human when the robot approached a waypoint that it was confused about.
Normally the band was kept uninflated ($0$ psi), but when the robot moved close to a waypoint where it was uncertain, the band inflated ($3$ psi) to squeeze the human's wrist and alert them about the robot's confusion.
In parallel to the haptic wristband, a wearable augmented reality headset overlaid visual markers on the robot's environment.
Users could see these markers as they interacted with the robot.
The markers corresponded to the robot's waypoints (also shown in \textbf{GUI}), but now contextualized those waypoints within the robot's physical environment.
Our combined \textbf{AR+Haptic} interface is shown in Figure~\ref{fig:user}.
By looking through the AR headset participants could identify i) the waypoints the robot had learned, and by wearing the haptic wristband participants were notified ii) when the robot learner was confident or confused.

\begin{figure*}
\centering
\includegraphics[width=2\columnwidth]{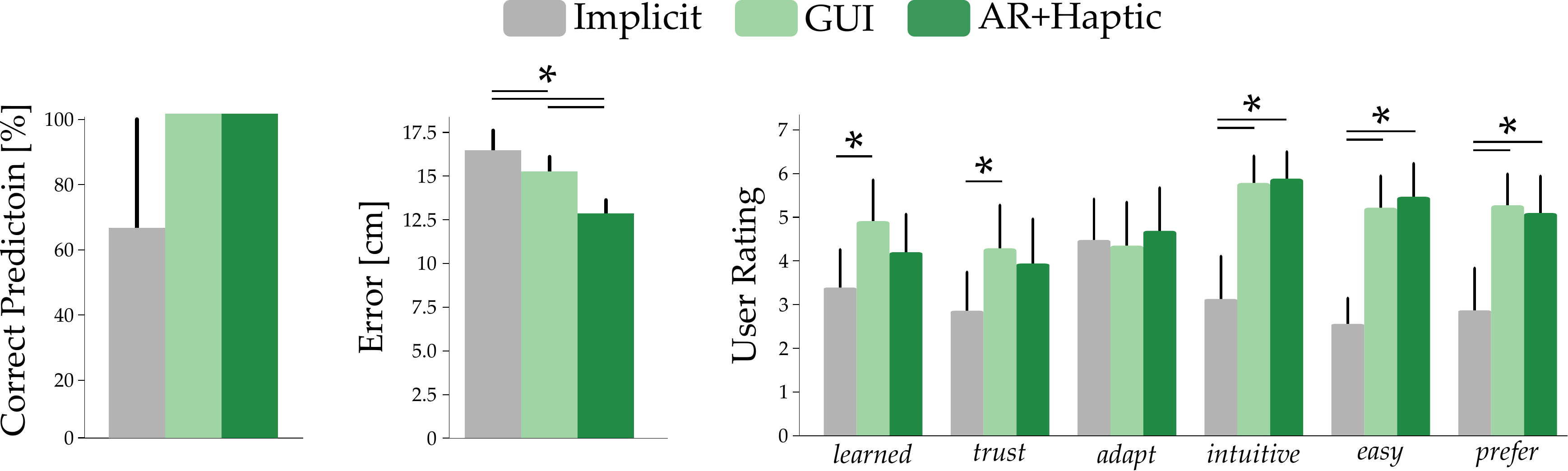}
    \caption{Objective and subjective results from our case study in Section~\ref{sec:userstudy}. Participants physically taught a robot arm by correcting its waypoints. The system closed the loop on robot learning in three different ways: i) including the human in the learning process without any additional signals (\textbf{Implicit}), ii) displaying the robot's learning on a computer monitor (\textbf{GUI}), or iii) rending the robot's learning across a wearable augmented reality headset and haptic wristband (\textbf{AR+Haptic}). (Left) When using \textbf{GUI} and \textbf{AR+Haptic} users could always tell which chair leg the robot was trying to assemble and align their teaching with this intended subtask. (Middle) As a result, during \textbf{GUI} and \textbf{AR+Haptic} the robot learned to assemble the chair legs with less error. (Right) When surveyed, users perceived the robots that explicitly communicated their learning (\textbf{GUI} and \textbf{AR+Haptic}) as better learners, more trustworthy, more intuitive to work with, and easier to understand. The participants preferred \textbf{GUI} and \textbf{AR+Haptic} overall as compared to \textbf{Implicit} feedback. Error bars show SEM and an $*$ denotes statistical significance ($p < 0.05$).} 
\label{fig:user_result}
\end{figure*}

\p{Dependent Measures} We used two objective metrics to analyze the outcomes of closing the loop on robot learning (Section~\ref{sec:H2}). 
We first measured \textit{Correct Prediction}, which records how frequently users were able to correctly predict which chair leg the robot was assembling. 
A higher \textit{Correct Prediction} percentage indicates that the human understood the robot's behavior and provided corrections for the same subtask the robot was trying to perform. Hence, \textit{Correct Prediction} was a measure of the human's teaching quality.
To assess the improvement in robot learning, we also measured the robot's \textit{Error} in assembling the chair legs after receiving corrections from the human.
We calculated \textit{Error} by measuring the total distance between the robot's end-effector and the actual locations of the chair parts before and after interacting with the participant. 
A lower \textit{Error} reveals that the participant's teaching more accurately conveyed how the robot should assemble the chair.

In addition to these two objective metrics, we also applied subjective questionnaires to probe the human's mental model of the robot learner (Section~\ref{sec:H1}).
Participants responded to the 7-point Likert scale survey shown in Table~\ref{table:likert} after interacting with \textbf{Implicit}, \textbf{GUI}, and \textbf{AR+Haptic}. 
This survey was composed using the practices recommended by \cite{schrum2023concerning}.
We included five multi-item scales and one single-item scale. 
Our survey asked participants whether it seemed like the robot \textit{learned} from their inputs, if they \textit{trusted} the robot, whether they \textit{adapted} their teaching over time, how \textit{intuitive} it was for them to teach the robot, how \textit{easy} it was to understand the robot's feedback, and to what extent they \textit{preferred} that communication method.

\p{Hypothesis}
We had two hypotheses for this case study:
\begin{displayquote}
    \textbf{H1.} \emph{Explicit feedback (\textbf{GUI} and \textbf{AR+Haptic}) will cause the robot to learn to perform the desired task more accurately.}
\end{displayquote}
\begin{displayquote}
    \textbf{H2.} \emph{Participants will prefer teaching a robot with explicit \textbf{GUI} or \textbf{AR+Haptic} feedback as compared to \textbf{Implicit} feedback.}     
\end{displayquote}

\p{Participants and Procedure} We recruited $12$ participants ($5$ female, $7$ male, average age $25$, age range $21$ -- $35$ years) from the Virginia Tech community. All participants provided informed written consent consistent with university guidelines (IRB \#$20$-$755$). Eight of the participants reported that they had interacted with a robot before. 

The experiment started with a familiarization procedure.
Participants practiced physically interacting with the robot arm and using each communication interface.
This familiarization phase lasted for a maximum of $15$ minutes.
Prior to the experiment the robot was pre-trained with an initial understanding of the chair assembly task; we provided a few expert datapoints offline, and then used this small dataset to train the robot's initial policy.
However, because the initial dataset was insufficient to convey the entire task, participants needed to intervene and continue teaching the robot during the experiment.
We followed a within-subjects design: each user completed the chair assembling task with \textbf{Implicit}, \textbf{GUI}, and \textbf{AR+Haptic}.
The order of these conditions was counterbalanced, where four users started with \textbf{Implicit}, four users started with \textbf{GUI}, and the final four users started with \textbf{AR+Haptic}.
During each trial participants observed the robot's behaviors and feedback and were free to intervene and correct the robot whenever they chose.
The robot recorded the user's expert inputs, re-trained its policy, and updated the feedback throughout the task.
Between each method the robot reset its learning so that it always started with an incomplete understanding of the task.

\subsection{Results and Discussion}

\p{Objective Measures}
The results for \textit{Correct Prediction} and \textit{Error} are displayed in Figure~\ref{fig:user_result}.

When participants received explicit feedback about the robot's learning they were able to more accurately predict what chair leg the robot was trying to assemble (\textit{Correct Prediction}).
With both \textbf{GUI} and \textbf{AR+Haptic} users intervened and provided expert datapoints for the subtask the robot was actually performing in $100\%$ of trials. 
By contrast, when users had only \textbf{Implicit} feedback, they misinterpreted the robot's intended behavior $35\%$ of the time, resulting in participants teaching the robot about a subtask it was not trying to perform.
For example, in Figure~\ref{fig:user} the \textbf{Implicit} robot was originally trying reach for the left-most chair leg, but the user instead ``corrected'' this motion by moving the robot to the right-most chair leg.

This misaligned human teaching caused the robot learner to have larger \textit{Errors} between where it should have reached for and placed the chair legs and where it actually moved.
A repeated measures ANOVA showed that different methods for communicating learning had a significant effect on \textit{Error} ($p<.001$).
Post hoc tests indicated that both \textbf{GUI} ($p<.05$) and \textbf{AR+Haptic} ($p<.001$) interfaces outperformed the \textbf{Implicit} condition. Comparing these two explicit methods, we found that \textbf{AR+Haptic} led to lower \textit{Error} than \textbf{GUI} ($p<.001$).
These results suggest that explicitly closing the loop and communicating the robot's learning to the human operator improves both the human's teaching (\textit{Correct Prediction}) and the robot's learning (\textit{Error}).

\p{Subjective Measures}
The results of the Likert scale survey are displayed in Table~\ref{table:likert} and Figure~\ref{fig:user_result}. 

We first checked the reliability of our five multi-item scales using Cronbach's $\alpha$ (where $\alpha>0.7$ was considered reliable).
\textit{Learned}, \textit{Trust}, \textit{Intuitive}, and \textit{Easy} were all found to be reliable scales, while \textit{Adapt} was not. The reliability of \textit{Prefer} was not tested since only one item (i.e., one question) was included on this scale. 
We next grouped each of the reliable multi-item scales into combined scores, and performed measures ANOVAs on each score.
The robot's method for communicating its learning had a significant main effect for \textit{Learned}, \textit{Trust}, \textit{Intuitive}, \textit{Easy}, and \textit{Prefer}.
To interpret these results and understand how \textbf{Implicit}, \textbf{GUI}, and \textbf{AR+Haptic} compared to one another, we finally applied the post hoc tests described below.

Starting with the \textit{Learned} results, participants thought that the robot learned what they wanted it to do more seamlessly with \textbf{GUI} than with \textbf{Implicit} ($p<.05$).
Similarly, users indicated that they \textit{Trusted} a robot with GUI feedback more than a robot with \textit{Implicit} feedback ($p<.05$). For both \textit{Learned} and \textit{Trust} the differences between \textbf{AR+Haptic} and \textbf{Implicit} were not statistically significant, although the average scores with \textbf{AR+Haptic} were higher.

The visual feedback from \textbf{GUI} and the immersive, multi-modal feedback from \textbf{AR+Haptic} made it clear to users what the robot had learned and what the robot was confused about. 
Participants indicated that explicit feedback communicated the robot's learning more \textit{Intuitively}, with both \textbf{AR+Haptic} and \textbf{GUI} getting higher scores than \textbf{Implicit} ($p<.01$).
Users also found this explicit feedback easier to interpret. Our post hoc tests showed that \textbf{GUI} and \textbf{AR+Haptic} were \textit{Easier} for participants to parse than \textbf{Implicit} ($p<.01$).
Overall, the users expressed a preference for explicit feedback interfaces over the \textbf{Implicit} condition.
When asked which method they preferred, the scores for \textbf{GUI} and \textbf{AR+Haptic} were both higher than \textbf{Implicit} ($p<.05$).

The only subjective result that did not follow this general trend focused on how humans \textit{Adapt} to the robot. 
When surveyed, users did not indicate that they adapted their teaching behavior in one condition any more than in another condition ($p=.70$). 
We note that this scale was also not reliable, and so participants may not have interpreted our survey questions about \textit{Adapt} in a consistent manner.

\p{Discussion}
In this case study we tested some of the trends across recent work on robot learning and communication.
We implemented a human-in-the-loop learning algorithm, which implicitly communicated what the robot had learned through its interactions with the human teacher. 
We extended that algorithm to extract explicit signals about the robot's learning, and then communicated those explicit signals via feedback interfaces.
These included a GUI that displayed learned waypoints on a $2$D computer monitor, as well as a multi-modal AR and haptic interface that rendered icons in the task environment and provided tactile alerts.
To assess the outcomes of closing the loop and communicating robot learning, we applied both objective performance metrics and subjective questionnaires about the human's experience.

Our results suggest that --- when the robot explicitly communicated what it was learning --- the human was better able to teach the desired task. 
We found support for improved human teaching in \textit{Correct Prediction}, which measured how frequently the human's teaching was aligned with the subtask the robot was attempting to perform.
This improved human teaching likely led to more accurate robot learning.
From \textit{Error}, we observed that robots which rendered explicit feedback learned to pick up and place the chair legs more precisely than robots which only provided implicit feedback.
Participants also perceived robots with explicit feedback as better learners: when asked if the robot \textit{Learned} what the human wanted, subjective scores were significantly higher for GUI as compared to implicit feedback.
These results support hypothesis \textbf{H1} and are in line with the ``Improved Teaching'' outcome from Section~\ref{sec:H2}.

Our results also suggest that communicating robot learning improves the human's perception of the interaction.
We found that the participants \textit{Trusted} the robot learner more when it provided explicit feedback about its learning. 
This could be because users felt they understood what the robot had learned and what the robot was confused about (\textit{Intuitive}), or because the explicit feedback was \textit{Easier} to interpret than the implicit human-in-the-loop approach.
One user wrote that \textbf{AR+Haptic} \textit{``communicated to me what the robot was trying to do and where it wanted to be.''}
Another stated that `\textit{`\textbf{GUI} helped me see exactly when to correct the robot.''}
By comparison, participants \textit{``did not like \textbf{Implicit} because it was not clear what was the robot's intention.''}
Overall, robots that conveyed what they had learned via explicit signals were subjectively \textit{Preferred} to the implicit approach.
These findings support \textbf{H2} and are in line with the ``Increased Trust'' outcome from Section~\ref{sec:H2}.

Our objective and subjective results suggest that explicit GUI or AR+Haptic feedback benefits the human-robot team.
But when comparing the visual GUI to the multi-modal AR+Haptic interface, the results were less clear cut.
On the one hand we found that AR+Haptic resulted in the least \textit{Error} in the robot learner, while on the other hand subjects perceived GUI as their most \textit{Preferred} method.
One advantage that the participants listed for AR+Haptic was their ability to focus on the task.
For instance, one user wrote: \textit{``GUI was distracting because I had to take my eyes off of the robot to watch the screen,''} and another said that \textit{``AR is better because I could see the robot's feedback from multiple perspectives.''}
Despite this issue, several users indicated that they still preferred GUI because the AR headset was heavy and uncomfortable to wear.
This highlights the importance of developing interfaces specifically for communicating robot learning, and making sure that these interfaces are user-friendly (see Section~\ref{sec:p2})

\section{Conclusion}

We have presented a cross-cutting review of communicating robot learning during human-robot interaction.
We surveyed three interdisciplinary branches of research: 
i) learning algorithms that determine what will be communicated, 
ii) communication interfaces that determine how communication will occur, and
iii) closed-loop measurements and outcomes that determine what constitutes effective communication.
When viewed together, this body of work covers how robots can close the loop between learning and communication.

Within each research area we identified underlying trends.
First, we found that robot learners can facilitate communication implicitly and explicitly. 
Implicit communication naturally arises when the human is involved throughout the learning process; in explicit frameworks the robot learner goes one step further to intentionally extract human-friendly signals that summarize what it has learned.
Second, we found that interfaces designed to communicate the robot's explicit signals are increasingly moving from computer screens to immersive systems.
This includes a transition from visual to non-visual feedback, as well as the emergence of multi-modal interfaces.
Finally, we found the tools applied to measure the effects of closing the loop, as well as the commonly measured outcomes.
A combination of objective metrics and subjective questionnaires are used to probe the human's model of the robot learner.
When robots communicate their learning, outcomes often include improved human teaching, increased trust, and co-adaptation.

These trends in research are driven in part by the still unsolved challenge of communicating robot learning back to human teachers.
To address this knowledge gap between what the robot has learned and what the human thinks the robot has learned, we proposed a series of open questions and interdisciplinary recommendations for future research. 
Our paper also included a new case study to illustrate how current robots can communicate their learning. 
This case study showed the positive interactions of the observed trends: our results indicated that explicit communication improves reported measures of learning, trust, and intuitiveness, and that immersive multi-modal interfaces allow better situational awareness for the human, leading to robots that more accurately learn the desired task.

\subsection*{Funding}

This work is supported in part by NSF Grants $\#2129201$ and $\#2129155$ and by the NSF Graduate Research Fellowship Program (NSF GRFP).

%%%%%%%%%%%%%%%%%%%%%%%%%%%%%%%%%%%%%%%%%%%%%%%%%%%%%%%%%%%%%%%%%%%%%%%%%%%%%%%%%%%%%%%%

\balance
\bibliographystyle{SageH}
\bibliography{references.bib}

%%%%%%%%%%%%%%%%%%%%%%%%%%%%%%%%%%%%%%%%%%%%%%%%%%%%%%%%%%%%%%%%%%%%%%%%%%%%%%%%%%%%%%%%

\end{document}